\documentclass[10pt,twocolumn,letterpaper]{article}

\usepackage[pagenumbers]{cvpr} 
\usepackage{cuted} 
\usepackage{caption}

\definecolor{cvprblue}{rgb}{0.21,0.49,0.74}
\usepackage[pagebackref,breaklinks,colorlinks,allcolors=cvprblue]{hyperref}
\usepackage{booktabs}
\usepackage{siunitx}
\usepackage{multirow}
\usepackage{array}
\usepackage{adjustbox}
\usepackage{float}
\usepackage{latexml}

\title{Active Vision Transformers}

\newcommand{\name}{\textit{aViT}}

\iflatexml
    \usepackage{authblk}

    \author[1]{Nicholas Fry}
    \author[1]{Shinjeong Kim}
    \author[2]{Maciej Lewandowski}
    \author[1]{Marwan Taher}
    \author[2]{Piotr Dudek}
    \author[1]{Paul H.~J.~Kelly}
    \author[1]{Andrew J.~Davison}

    \affil[1]{Imperial College London}
    \affil[2]{University of Manchester}
\else
    \author{
    Nicholas Fry$^{1}$\thanks{Equal contribution.} \quad
    Shinjeong Kim$^{1*}$ \quad
    Maciej Lewandowski$^{2}$ \quad
    Marwan Taher$^{1}$ \\
    Piotr Dudek$^{2}$ \quad
    Paul H.~J.~Kelly$^{1}$ \quad
    Andrew J.~Davison$^{1}$ \\[4pt]
    $^{1}$Imperial College London \qquad
    $^{2}$University of Manchester
    }
\fi

\begin{document}

\maketitle
\iflatexml
    \begin{center}
        \small
        Nicholas Fry and Shinjeong Kim contributed equally.
    \end{center}
\fi

\begin{strip}
  \centering
  \vspace*{-46pt}
  \captionsetup{type=figure}
  {\includegraphics[width=\linewidth,
    trim=0 3 0 0, clip]{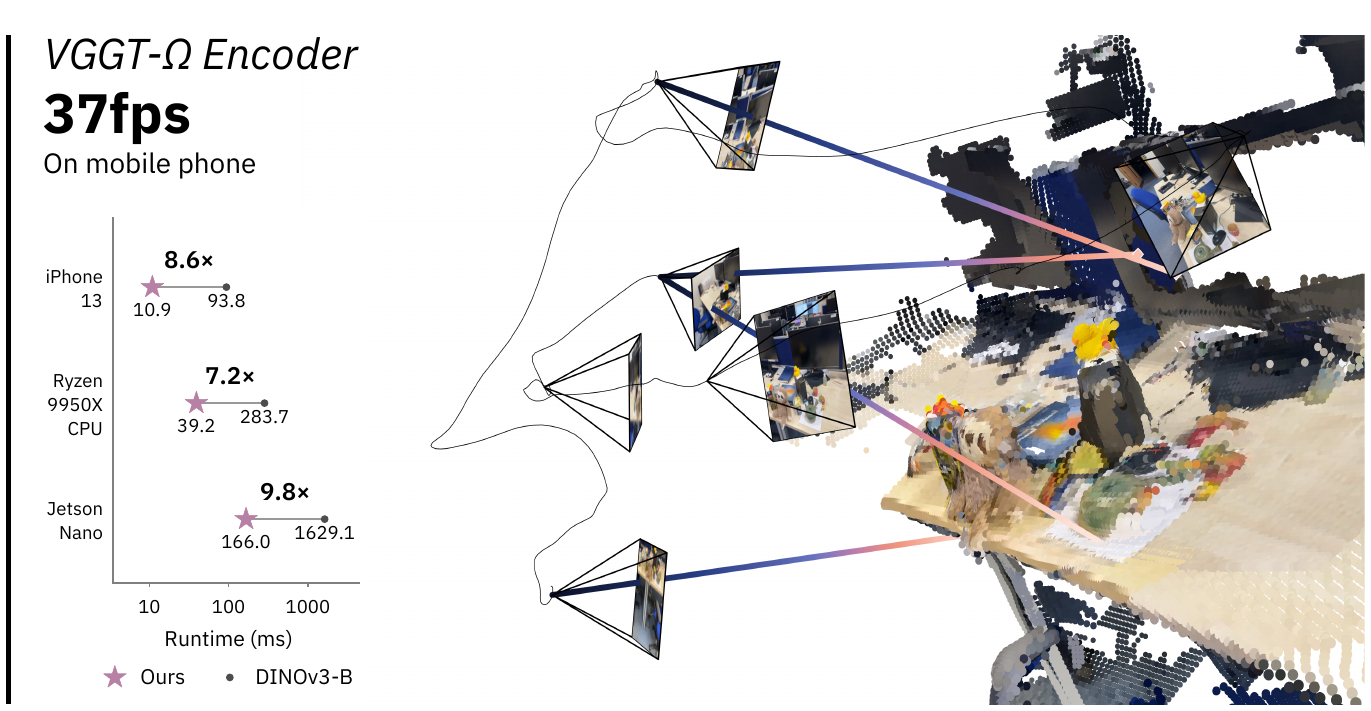}}
  \setlength{\abovecaptionskip}{1pt}
  \vspace{-2mm}
  \captionof{figure}{\textbf{Real-time dense feature encoding.}
\emph{Left:} runtime (log scale) of one \name{}-B glimpse and memory update with a DINOv3 ViT-B encoder (stars), against DINOv3 ViT-B on the whole frame, on an iPhone 13, a 16-core desktop CPU (Ryzen 9950X) and a Jetson Nano. \name{}s run at up to 92\,fps on the phone.
\emph{Right:} \name{}-L encoder of VGGT-Omega~\cite{wang2026vggtomega}, running live at 37\,fps on the phone. Each frame is encoded with two glimpses, and the features are sent to a PC, where KV-Tracker~\cite{Taher_2026_CVPR} decodes the 3D map and poses.}
  \label{fig:teaser}
\end{strip}
\begin{abstract}
To meet the tight latency and power constraints of edge devices, we propose Active Vision Transformers (\name{}s), an architecture that can adapt pre-trained task-agnostic vision foundation models, such as DINO, to be more compute efficient via foveated tokenisation. We add a persistent lightweight spatial memory to accumulate dense scene features over time and show \name{}s can accurately capture the features from a range of different teacher models. To our knowledge, we outperform all other active vision foundation models on 2D image tasks such as segmentation and depth prediction, with comparable/fewer FLOPs and only 22 hours of training on a single GPU. 
We show that an \name{} can achieve results that are competitive with a full ViT-B/L encoder while being up to 10$\times$ computationally cheaper. We demonstrate our method with VGGT-Omega's encoder running in \textbf{real-time 37fps} (iPhone 13). Project Page: \url{https://www.shinjeongkim.com/avit/}
\vspace{-5mm}
\end{abstract}
    
\section{Introduction}
\label{sec:intro}
Power and latency are critical for robots and wearable devices.
Embedded hardware has therefore favoured convolutional neural networks, in which each kernel reads only a small neighbourhood, such that computation and data movement stay cheap~\cite{sze2017, energy_problem_horowitz}.
Vision Transformers (ViT) have shown that aggregating information across the whole image plane with attention gives higher accuracy than convolutions on most downstream tasks ~\cite{Kolesnikov2021,oquab2024dinov2}. 

However, 
attention is slow and costly in energy on embedded devices, because every token is compared with every other.
Its cost grows as $O(N^2)$ in the number of tokens $N$, and data must travel long physical distances and persist over long time spans.
Parallel hardware reduces the latency of attention but not its energy.
The number of operations is unchanged, and running more compute units at once draws more power, which is the limiting resource on an edge device~\cite{energy_problem_horowitz}. 

Many methods reduce this cost by changing the attention operation itself.
Linear attention methods approximate it at a cost linear in $N$, and FlashAttention~\cite{flashAttentionDao} computes it exactly with less memory traffic.
However, these methods do not depend on the content of the image.
They make each token cheaper to process, but they still process every token, even if the information extracted is not needed.

The alternative is to reduce $N$, by reading from the scene only the information that the task requires.
In biological vision, humans and animals allocate more cognitive capacity to the parts of a scene that are complex or changing over time, and less to parts that are simple and static. 
This is most obvious in the continual  re-direction of gaze, and the eye's high resolution fovea, towards scene elements of current interest.
There is a long history in computer vision of Active Vision approaches to apply the same idea to machines~\cite{Aloimonos1988,activeFixation,Davison:Murray:ECCV1998, Murray:etal:ICCV93}. An active vision system controls what it observes, and chooses each observation to suit its task, in contrast to the blanket processing. Active control can be of the physical direction of a robotic camera, or of the focusing of processing resources within images captured passively~\cite{Davison:ICCV2003}.
A variety of dynamic tokenisation methods have brought this principle to modern ViTs. Approaches that aggregate information sequentially, starting from regions of interest, regions of high scene complexity, or regions of high information gain, have been explored across many tasks~\cite{Shi_2026_CVPR, nakayama2026leveloftokendiffusion, 2026eyerobot2}. 
However, most of these methods are designed to produce the output for one predefined task, and few aim for general features that can serve many tasks.
Their architectures are also complex, which either increases inference time or makes them generalise poorly across gaze policies.
Foveation is the exception.
It is conditioned only on a point of interest, an input that many tasks and interfaces already provide.

In this paper, we introduce Active Vision Transformers (\name{}s), a general architecture that can convert off-the-shelf pretrained ViTs into active vision models, which build up dense features of a scene from foveated glimpses chosen by a policy. Each glimpse captures detail at the query point and context from its peripheral vision. A memory aggregates information across glimpses, so detail accumulates if the task requires it. Human vision works in a similar way: we see sharply only where we look, and build our impression of a whole scene from moving our fixation \cite{carrasco1995eccentricity}.

Our contributions are as follows:
\begin{itemize}
\item We propose a general ViT-based architecture for active vision foundation models with a memory module that propagates information across glimpses and across frames in video sequences. 
\item We introduce a dynamic tokenisation scheme whose representation capacity can adapt to a compute budget. The scheme supports zooming and fixation at multiple points, with any number of fixations per frame. The number of tokens, and hence the cost of each glimpse, can therefore be set to meet a given latency budget.
\item We show state-of-the-art performance across a number of tasks including depth estimation, semantic segmentation, video segmentation, etc. We also benchmark our runtime performance on different processors and find that we improve runtime by up to 10$\times$ compared to the baseline architectures. 
\end{itemize}

Active ViTs enable real-time dense feature extraction on embedded devices. We use our architecture to adapt the encoder of VGGT-Omega~\cite{wang2026vggtomega}, a large scale Vision Transformer (ViT-L), and run it at \textbf{37\,fps on an iPhone 13}, compared with 3\,fps for the original encoder. We demonstrate that  DINOv3 ViT-B is \textbf{7 to 10 times faster} than the original encoder on a phone, a 16-core desktop CPU and a Jetson Nano, reaching up to \textbf{92\,fps} on the phone (\cref{fig:teaser}).
\section{Related work}
\label{sec:related}
\paragraph{Reading less of the image.}
The cost of a vision transformer grows with the number of tokens it reads, so a direct way to save computation is to read fewer.
Recurrent attention models classify an image from a sequence of small glimpses~\cite{mnih2014ram,ba2015multiple,wang2020glance}. These models are trained for classification and summarise their glimpses in a single vector, so they cannot be used for tasks requiring dense or more general features.
For vision transformers, token pruning and merging remove tokens inside the network~\cite{rao2021dynamicvit,bolya2023token}, but only after the whole image has been read and embedded.

Segment This Thing (STT)~\cite{schmidt2025stt} reduces the input by sampling larger patches further from a query point and downsampling each to the same size.
It segments the clicked object from these foveated tokens, which are fine at the click and coarse at the edge of the field. Foveated inputs have since been used to guide robot policies~\cite{wetzstein2026policy, chen2026foveavla, 2026eyerobot2}.
STT, however, is trained solely for single object segmentation. From a glimpse, it can describe segmentation masks for the user query, but does not generalise further.
We extend STT's sampling pattern, and use it as the input to a general encoder with a memory of the scene.

\paragraph{Accumulating partial observations.}
A robot navigating a room needs only coarse estimates of the whole scene, while one that grasps or tracks a fast
object needs fine detail where the object is~\cite{2026eyerobot2}.
Lowering the resolution or cropping the input fixes this trade-off in advance~\cite{thavamani2021fovea}.
Foveation lets the fixation decide it instead.
Holding the fixation on one object resolves that object in detail on every frame.
Moving it across the scene builds up a view of the whole, which requires a memory of earlier glimpses.
Active visual exploration methods combine glimpses of a still image to recover its
content~\cite{seifi2021glimpse,pardyl2023active,pardyl2024adaglimpse}.
 
The closest work to ours, CanViT~\cite{berreby2026canvit}, writes each glimpse into a \emph{canvas}, a grid of dense features aligned with the scene dimension and trained to match DINOv3~\cite{simeoni2026dinov3} features.
It is trained from scratch on still images.
In video, earlier work reuses computation over time instead, by propagating the features of key frames~\cite{zhu2017deep}
or updating only the tokens or pixels that change~\cite{dutson2023eventful,parger2022deltacnn}, but these methods still
take the whole frame as input.
Recurrent transformer models \cite{dehghani2019universaltransformers} such as CUT3R~\cite{wang2025cut3r}, D\'ej\`a View \cite{burzio2026dejaview} and Solv3R~\cite{kwak2026solv3r} carry one scene
state across views.
Our state is also carried from step to step, but it is a grid of spatially  aligned features updated via attention.

\paragraph{Efficient visual foundational models.}
Since visual foundation models (VFMs) rose to prominence for their cross-domain and cross-task generalisation~\cite{radford2021learning,oquab2024dinov2,kirillov2023segment,simeoni2026dinov3}, there have been continued efforts to make them lighter and thus better suited to constrained settings such as edge applications. These include methods that build lightweight backbone through dynamic tokenisation, as discussed above~\cite{rao2021dynamicvit, bolya2023token,schmidt2025stt,berreby2026canvit}, or through quantisation~\cite{yuan2022ptq4vit,li2023repq}. Among them, our work is influenced by methods that exploit the locality of image data, either by concentrating computation on a subset of mutually relevant information~\cite{wang2020glance,schmidt2025stt} or by aggregating such partial results along the spatial~\cite{seifi2021glimpse,pardyl2023active,pardyl2024adaglimpse,berreby2026canvit} or temporal~\cite{zhu2017deep,dutson2023eventful} axis to recover the full output.

Token merging~\cite{bolya2023token} and post-training quantisation~\cite{yuan2022ptq4vit,li2023repq} need no retraining of the backbone.
Co-Me~\cite{chen2026comeconfidenceguidedtokenmerging} merges the tokens that a small distilled predictor marks as low in confidence, and reports speed-ups of up to $21.5\times$ for VGGT \cite{wang2025vggt} on sequences of 512 frames.
However, merging starts only after the middle of the encoder, and the speed-up falls to $1.13\times$ for a single frame.
These methods therefore still read and embed the entire input.

Aggregating partial observations into foundation model features avoids this but requires training from scratch~\cite{berreby2026canvit}. We observe that teacher-student distillation has been performed with a single~\cite{oquab2024dinov2,simeoni2026dinov3,zhang2025accessing} or multiple~\cite{ranzinger2024amradio,sariyildiz2024unic,shang2024theia} teacher VFMs. Among these approaches, we are particularly inspired by methods that initialise the student by inheriting weights from any teacher of the same architecture family~\cite{wu2023tinyclip,xu2024initializing}. However, none of them aggregates partial observations, and the active vision model distilled from a VFM~\cite{berreby2026canvit} does not inherit its weights.\vspace{3mm}

To our knowledge, we introduce the first active vision method that reduces the cost of foundation models without being tied to a particular ViT backbone.
As a result, edge devices can obtain features close to those of any ViT-based ~\cite{simeoni2026dinov3} architecture at a tenth of the cost, for any spatial intelligence task they may be required to solve.\vspace{-2mm}

\paragraph{Processing at the sensor.}
Moving data costs more energy than computing on it~\cite{energy_problem_horowitz}, so efficient vision systems keep data close to where it is produced~\cite{fry2026}. The cheapest pixels are those that never leave the sensor, and some sensors already select what they send. Event cameras report only the pixels whose brightness changes~\cite{gallego2020event}. Pixel processor arrays go further and place a small processor next to every photodiode~\cite{carey2013scamp5}, so that a neural network can run on the sensor itself~\cite{bose2019camera}. On this hardware, the recurrent network PixelRNN sends up to 256 times less data than the raw sensor output~\cite{so2024pixelrnn}, and a programmable silicon retina sends filtered, retina-like events instead of frames~\cite{lewandowski2026retina}. In each case, the output is reduced by a rule fixed in advance, such as a change threshold or a network trained for one task. Foveation instead lets a policy decide which pixels leave the sensor, and the policy can change with the task. A pixel processor array could read out only the foveated pixels that an \name{} takes as input, which is a step towards running ViT foundation models on very low-power devices.

\section{Method}
\label{sec:method}

\subsection{Overview}
\label{sec:overview}
Our method can convert off-the-shelf pretrained vision transformers (ViTs) into active vision transformers (\name{}s). The architecture reads an image through foveated glimpses, and writes each glimpse into a persistent memory, where dense features build up over successive glimpses.
Our goal is a ViT that fits a small compute budget on an embedded device.
It should spend more of its capacity on regions of interest, and refine its dense features by moving its fixation to other parts of the image as needed.
We achieve this by taking a pretrained ViT model, which also serves as the teacher, and initialise a matching encoder from its weights.
The encoder's output is fed into a latent memory module, which is updated at every glimpse.
We refer to this memory as the \emph{canvas}~\cite{berreby2026canvit}.
A glimpse is a set of tokens sampled around a fixation point, with fine patches around the fixation point and coarse further away from it. 

\begin{figure*}[t]
    \centering
   \vspace{-4mm}
    
   \includegraphics[width=0.9\linewidth, trim=0 1.7cm 0 2cm, clip]{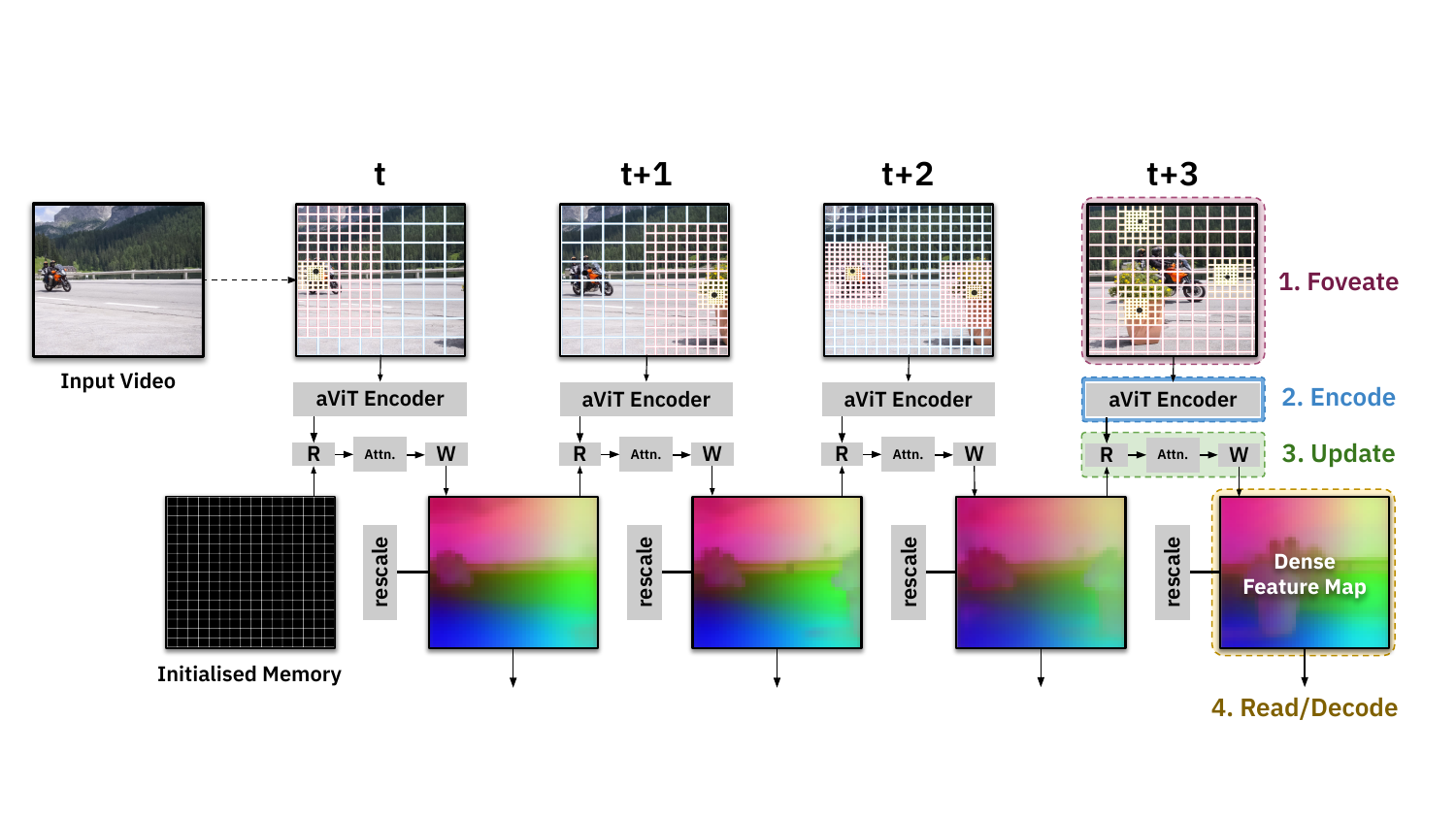}
    \caption{ Inference of an Active Vision Transformer (\name{}) on a video sequence.}
    
    \label{fig:method}
\end{figure*}

A forward pass of an \name{} goes through three main stages (\cref{fig:method}).
First, the frame is \emph{foveated} into about 100 tokens (\cref{sec:fovea}).
Second, the fine-tuned backbone \emph{encodes} the tokens, without seeing the canvas (\cref{sec:encoder}).
Third, the tokens \emph{update} the canvas: they read from it, pass through two transformer layers and are written back to it (\cref{sec:canvas}). At any point the dense features may be read out and used for the downstream task.

Starting from DINOv3's weights, an \name{} needs only 22 hours of fine-tuning.
At run-time, its encoder costs 17.8\,GFLOPs for one glimpse of a 720\,px frame, on average over fixations,
against 499\,GFLOPs for DINOv3 on the whole frame, 28 times as much for the encoder alone and about 21 times with the canvas update (\cref{tab:multifovea_cost}). The loss in quality depends on the number of glimpses (\cref{fig:pentagon_pair}, \cref{sec:experiments}).
Several fixations can also be combined in one tiling of the frame (\cref{sec:multifovea}).
They share their coarse tokens, so the cost grows more slowly than the number of fixations.

\subsection{Foveation}
\label{sec:fovea}
Each glimpse of a frame starts from one or more fixation points chosen by an external policy (\cref{sec:experiments}).
For a single fixation $f$, we sample a square field of side $F = 1{,}280$\,px around it, using the pattern of Segment This Thing (STT)~\cite{schmidt2025stt}.
The pattern has 172 square boxes (patches) in five levels, with sides from 16\,px at the fixation to 128\,px at the edge of the field; details are shown in the supplementary material.
Each box is averaged down to $16 \times 16$ pixels \cite{Kolesnikov2021}, so every token holds the same number of values whatever area it covers. The 16\,px boxes at the fixation need no averaging, so the fovea is read at the native resolution of the frame.
Boxes that fall outside the frame are dropped and not passed to the encoder.
On a 720\,px frame a glimpse keeps 97 tokens on average (between 43 and 121, depending on the fixation), against the 2{,}025 patches that DINOv3 processes for the whole frame. Self-attention grows with the square of the number of tokens, but at this size it accounts for only 2\% of the
encoder's FLOPs, against 30\% for DINOv3 on the whole frame.
The cost of a glimpse is therefore close to linear in the number of tokens.

\paragraph{Multiple fixations.}
\label{sec:multifovea}
A glimpse need not be restricted to a single fixation.
A policy may choose both the number of fixations $K$ and their positions.
We combine the fixations in one tiling of the frame, and show  that our method generalises to this setting even when not being specifically trained for it (\cref{fig:multi-fovea-qual}).
To build the tiling, we assign a box side to every pixel $\mathbf{x}$ of the frame.
Let $d_k$ be the pixel distance from $\mathbf{x}$ to fixation $\mathbf{f}_k$, taken as the larger of the horizontal and
vertical offsets.
Fixation $k$ alone would assign a side of 16\,px if $d_k < 32$\,px, 32\,px if $d_k < 64$\,px, etc.
The pixel takes the smallest of these sides over all fixations.
The frame is first covered by $128 \times 128$\,px squares aligned with the frame, and each square is split into quarters
while any of its pixels has been assigned a side smaller than the square's.
Every pixel is then covered by exactly one token, with a side of 128, 64, 32 or 16\,px.
The fixations share their coarse tokens, so the number of tokens grows more slowly than $K$.
In a 512\,px scene, four fixations on one object give 155 tokens on average, against 115 for the tiling of one fixation.

\paragraph{Zooming.}
We also introduce a zoom factor $z$ to scale the foveation pattern. 
Let box $i$ have centre $p_i$ and side $s_i$ in frame pixels at $z = 1$.
At zoom $z$, the box has side $s_i / z$ and centre $f + (p_i - f)/z$.
A zoom $z > 1$ therefore gives finer tokens over a smaller field, and $z < 1$ gives coarser tokens over a wider one. See \cref{sec:experiments} for further details.

\paragraph{Glimpse policy.}
\name{}s are agnostic to the glimpse strategy. We train on sequences of foveae sampled uniformly at random over the image plane (U-IID), so that it can extract useful information from any glimpse strategy at test time, an approach inspired by CanViT.

\subsection{Encoder}
\label{sec:encoder}
The encoder is a ViT, that we initialise from the target teacher model weights.
We fine-tune it to be compatible with the tokens from our foveation pattern (\cref{sec:fovea}); its architecture is otherwise unchanged.
The method can therefore turn any pretrained ViT backbone into a foveated encoder with a scene memory. 
We demonstrate this with several pre-trained backbones in (\cref{sec:experiments}). The encoder is joinlty fine-tuned along with training the canvas memory module, in our experiements we show results with both full weight fine-tuning and LoRA (VGGT-Omega) ~\cite{hu2022lora}.

\subsection{The canvas}
\label{sec:canvas}
The canvas memory has the same shape as the teacher's dense feature output.
Given a frame of dimensions $H \times W$ pixels, the size of its corresponding feature map is a grid of $\tfrac{H}{16} \times \tfrac{W}{16}$ tokens of length $d$. Each token is the corresponding feature for a $16 \times 16$ pixel patch~\cite{Kolesnikov2021}.
We train the model such that each canvas token encodes the features that the teacher would predict for its corresponding full resolution patch. Each glimpse updates the canvas, while its content is propagated from one glimpse to the next and from one video frame to the next.
It therefore gains detail wherever the model has looked, and a downstream task can use it at any time in place of running the full foundation model encoder.
For a DINOv3 ViT-B/16 teacher model, the canvas has $45 \times 45$ cells of width 768 for a 720\,px scene, and $32 \times 32$ cells for a 512\,px scene.
At the start of a sequence, every cell takes learned initial values.
The read-out is a single LayerNorm, which turns each cell into the predicted feature for its patch.
 
\paragraph{Updating the canvas.}
After the foveated patches are encoded (\cref{sec:encoder}), they interact with canvas in the memory module updating the canvas tokens, which happens over four steps:
\begin{enumerate}
  \item \textbf{Read}: Foveated tokens are updated via a residual from cross-attention with the canvas tokens.
  \item \textbf{Self-attention block}: Foveated tokens are propagated through two self-attention layers.
  \item \textbf{Write}: Canvas cells are updated via a residual from cross attending to the glimpse tokens.
  \item \textbf{Rescale}: Canvas tokens are RMS normalised, to prevent the their values from growing after many writes, for example when the canvas is propagated across video frames.
\end{enumerate}

Following CanViT's canvas recipe we use SR-RoPE for canvas reads and writes, and with a LayerNorm to obtain key, queries and values from the canvas tokens, rather than using learned projections. 

\subsection{Position encoding}
\label{sec:rope}

Self-attention is permutation-equivariant,  so any spatial structure must be supplied through a position encoding or learned from the content of the tokens. DINOv3 uses RoPE~\cite{su2024roformer} to encode the relative token positions during self-attention between the patch tokens that share the same size and live on a fixed grid.
Since our model dynamically allocates patches of different sizes based on the fixations 2D location, we give each token the position of its box centre, measured from the fixation. RoPE depends only on differences in position, so the choice of origin does not change attention. The model can infer a token's size because the centres of larger boxes lie further apart. 
For our memory module we adopt SR-RoPE \cite{berreby2026canvit}, which defines the positions to be with respect to the centre of the scene normalised from $[-1, 1]$. This allows the encoder tokens and the canvas tokens to exchange relative spatial information while having different  spatial dimensions.

\begin{figure*}[t]
  \centering
  \includegraphics[width=\linewidth]{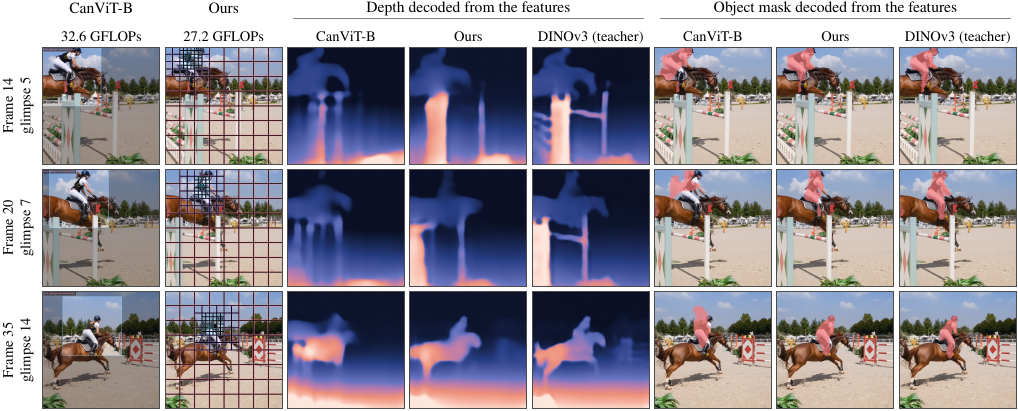}
  \caption{\textbf{An \name{}-B and CanViT-B on a video, with the canvas carried between frames.}
  Left: what each model reads on three frames, with its mean cost per frame.
  Right: depth and segmentation decoded from the features of CanViT-B, \name{}-B and the DINOv3 teacher.}
  \label{fig:video_fovit}
  \vspace{-2mm}
\end{figure*}

\subsection{Training recipe}
\label{sec:training}
Fine-tuning trains the encoder and the canvas together, so that the canvas reproduces what the teacher would output for the whole frame (\cref{sec:overview}).
The encoder starts from the teacher's weights, and the canvas and its update start from new weights.
We describe the target and the loss first, and then the sequences of glimpses used in training.

\paragraph{Target and loss.}
The target is the result of the teacher model run on the whole scene at full resolution, which gives one feature per $16 \times 16$ patch.
The canvas cells follow the same grid, so each cell has one target feature.
The target features are standardised per position and channel.
The loss is the mean squared distance between the read-out and the target,
\begin{equation}
  \mathcal{L} = \frac{1}{G N} \sum_{g=1}^{G} \sum_{i=1}^{N}
  \bigl\| \hat{\mathbf{y}}_{g,i} - \mathbf{y}_{g,i} \bigr\|^2 ,
\end{equation}
where $\hat{\mathbf{y}}_{g,i}$ is the read-out of cell $i$ after glimpse $g$, $\mathbf{y}_{g,i}$ its target, $G$ the number of trained glimpses and $N$ the number of cells.
Here $\|\cdot\|$ is the root mean square over the 768 (ViT-B) channels, so predicting the mean feature everywhere gives a loss of about 1. We supervise the canvas after every glimpse, not only the last, because a downstream task may read it at any time.
It should therefore be as accurate as the glimpses so far allow, and the model receives a training signal at every step of the sequence rather than only at the end. 

\paragraph{Training sequences.}
Half of the training steps use still scenes and half use video clips at 6 frames per second.
Each sequence begins without gradients, with up to 8 glimpses on a still or up to 4 frames of a clip, so that training also starts from partly filled canvases.
Stills then have 8 trained glimpses at random positions; each glimpse after the first stays at the previous fixation with probability 0.3.
Clips then have 6 trained frames of one glimpse each, at a new random position, with the canvas carried from frame to frame.
Consecutive frames are 0, 1 or 2 frames apart with probabilities 0.2, 0.4 and 0.4, where 0 repeats the frame.

\paragraph{Implementation details.}  
\label{sec:impl}
 
We use AdamW with $\beta = (0.9, 0.95)$ and a weight decay of 0.05.
The learning rate is $5 \times 10^{-4}$ for the canvas parts and $2 \times 10^{-5}$ for the DINOv3 encoder.
It warms up over 500 steps and then follows a cosine decay to a tenth of its peak.
We clip gradients at norm 1. The final models train for 100{,}000 steps with batches of 32 sequences, each on one GH200 Superchip.
For smaller experiments we train for 20{,}000 steps with batches of 8, each on one RTX 6000 Ada GPU.

\section{Experiments}
\label{sec:experiments}

\begin{figure}
    \centering
    \includegraphics[width=0.9\linewidth]{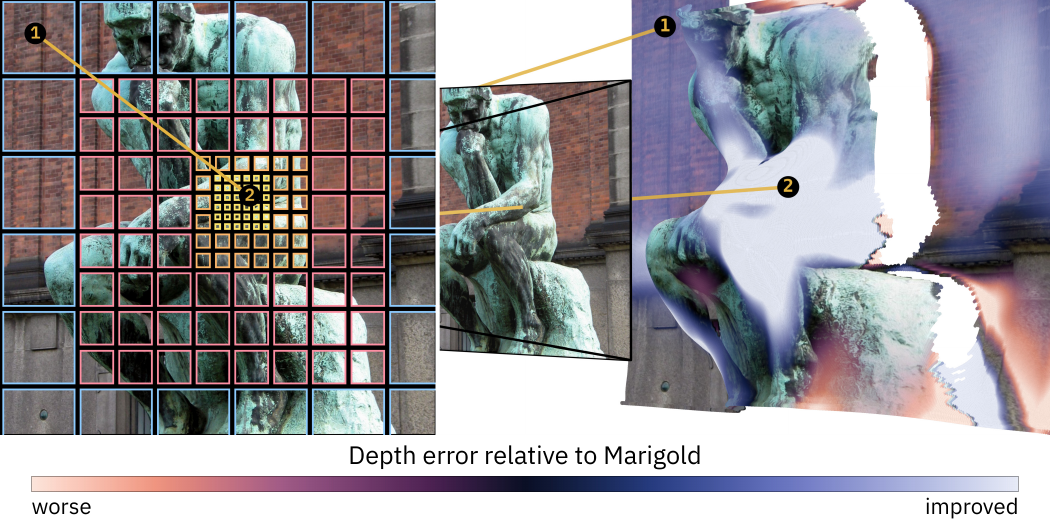}
    \caption{Improvement of depth estimation after moving fixation. We use Marigold \cite{pavlovic2026marigoldv2} as pseudo-ground truth. The reduction in RMSE, particularly around the fovea, demonstrates that an \name{} actively acquires information focused near points of interest.}
    \vspace{-3mm}
    \label{fig:depth_improvement}
\end{figure}

To ensure we evaluate the performance of our method fairly we also train CanViT-B~\cite{berreby2026canvit} with its own code, unchanged, on the same data, fixations, schedule and loss as our model, starting from both DINOv3's weights and from its released weights. We then evaluate on the release weights and the fine-tuned version and report the best performance.

\paragraph{Glimpse policies.}
\label{sec:glimpse_policies}
CanViT~\cite{berreby2026canvit} is trained with F-IID, which views the whole scene first and then samples glimpses at random positions and scales. Its best policy for video segmentation is EG-C2F, which visits the tiles of a quadtree from coarse to fine and, at each scale, picks the unvisited tile where a semantic segmentation probe is least certain, measured by the entropy of its class predictions.

We train \name{}s with U-IID, which places one fixation per glimpse at a uniformly random position. At test time, fixations can be placed where they are most useful, and objects are natural targets. Our object-based policies use a lightweight YOLO11 tracker~\cite{yolo} and sample each fixation from a normal distribution centred on an object, with a standard deviation of half the box size. The object-based policy can be matched to the latency budget. Under a tight budget, we place one fixation on one object (Obj). Under a looser budget, for example in a slowly changing scene, we sample several fixations and process them in one forward pass, either as a tiling (Obj-Tile) or as overlapping foveas (Obj-Overlap), as described in \cref{sec:multifovea}.
\vspace{-2mm}

\paragraph{Performance on video.}
\cref{fig:video_fovit} compares an \name{}-B with CanViT-B on a DAVIS~\cite{pont2017davis,Perazzi_2016_CVPR} clip that was not used in training, with each model's canvas carried between frames. On each frame, CanViT-B receives two glimpses, a view of the whole scene and then a crop around the query point, while the \name{} receives a single glimpse at the same point.
We decode depth and an object mask from the canvas of each model, and from the features of their shared teacher, DINOv3, run on the whole frame. All three use the same decoding heads, trained on DINOv3 features, so any difference in the output comes from the features.
The static image semantic segmentation head was trained on one frame from each YouTube-VOS 2019~\cite{xu2018youtubevos,Xu_2018_ECCV} training video and 1{,}000 SA-1B~\cite{kirillov2023sam} images, and the depth head was trained to match Depth Anything V2 Small~\cite{deptha2}.
\name{}s stay closer to the teacher on both tasks, at a lower cost per frame. 

\begin{figure}
    \centering
    \includegraphics[width=0.9\linewidth]{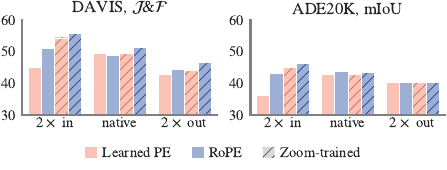}
\vspace{-3mm}
    \caption{Segmentation on 512\,px images with different zoom factors. DAVIS $\mathcal{J}\&\mathcal{F}$ and ADE20K
mIoU (8 glimpses a frame), for STT's learned position embedding and for RoPE, each trained with and without random zoom. Note that y axis starts at 30.}
    \label{fig:zoom_rope}
\end{figure}

\subsection{Generalization across tasks}
\begin{figure}[t]
    \centering
    \includegraphics[width=\linewidth]{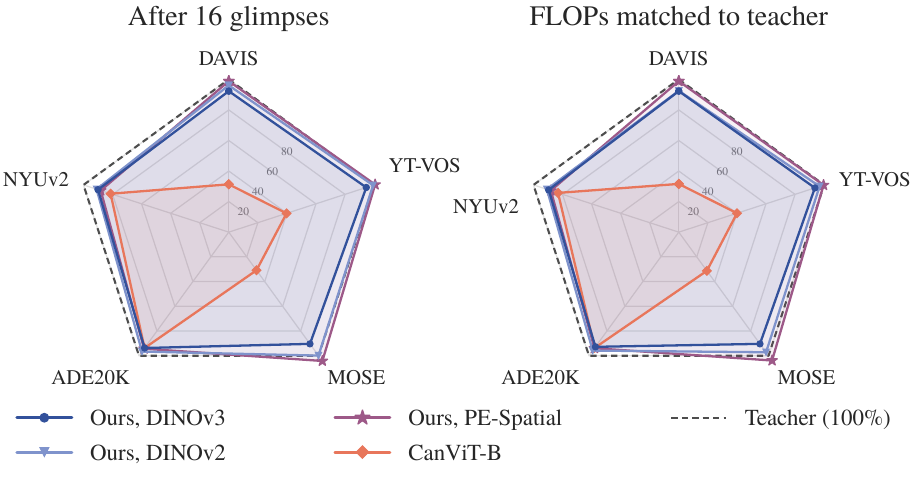}
    \caption{\textbf{Performance retention ratio} (the ratio of student performance to teacher performance) of our method with various models used as teachers. For metrics where lower values are better, such as Depth RMSE (NYUv2 dataset), the reciprocal of this ratio is used.}
    \label{fig:pentagon_pair}
    \vspace{-5mm}
\end{figure}

Our method can convert any ViT backbone with a short training phase. To show this, we distill \name{}s from several visual foundation models and evaluate each on video segmentation (DAVIS~\cite{pont2017davis,Perazzi_2016_CVPR}, YouTube-VOS~\cite{xu2018youtubevos,Xu_2018_ECCV} and MOSE~\cite{MOSE, MOSEv2}; $\mathcal{J}\&\mathcal{F}$), image segmentation (ADE20K~\cite{zhou2017scene, zhou2019semantic}; mIoU) and depth estimation (NYUv2~\cite{nyudataset}; RMSE).
For video segmentation, each model uses its best policy (EG-C2F for CanViT and Obj-Tile for the \name{}), and for the other tasks, the policy it was trained with (F-IID and U-IID; \cref{sec:glimpse_policies}).
Video segmentation is training-free, and following the evaluation protocol of DINOv3~\cite{simeoni2026dinov3}, we randomly sample our own evaluation splits at the same ratio, as theirs are not public.
In \cref{fig:pentagon_pair}, we measure how much of the teacher's performance each model retains. 
The left of the figure shows performance with 16 glimpses, and the right shows performance at the same computational cost as the teacher.
With a single fixation, at roughly a tenth of the teacher's FLOPs, \name{}s retain 70\% to 90\% of the teacher's performance, and over 90\% with 16 fixations (\cref{tab:fovit_canvit}).
\name{} surpasses the PE-Spatial teacher on MOSE. We suspect this is because Obj policy injects a prior on the object's location.

\paragraph{Training with zoom.} 
We evaluate whether RoPE helps the encoder generalise to zoomed glimpses (\cref{fig:zoom_rope}).
STT's learned slot embedding~\cite{schmidt2025stt} is the same at every zoom, so the encoder must infer each token's position and coverage from its pixels, and needs fine-tuning to handle zoom.
Trained without zoom, the encoder with RoPE generalises much better to zooming in, leading by 6.0 $\mathcal{J}\&\mathcal{F}$ on DAVIS and 7.0 mIoU on ADE20K at $2\times$.
At native resolution and at $2\times$ zoom out, the two encodings are within 1.3 points of each other.

\begin{figure*}[t]
    \centering
    \includegraphics[width=1\linewidth]{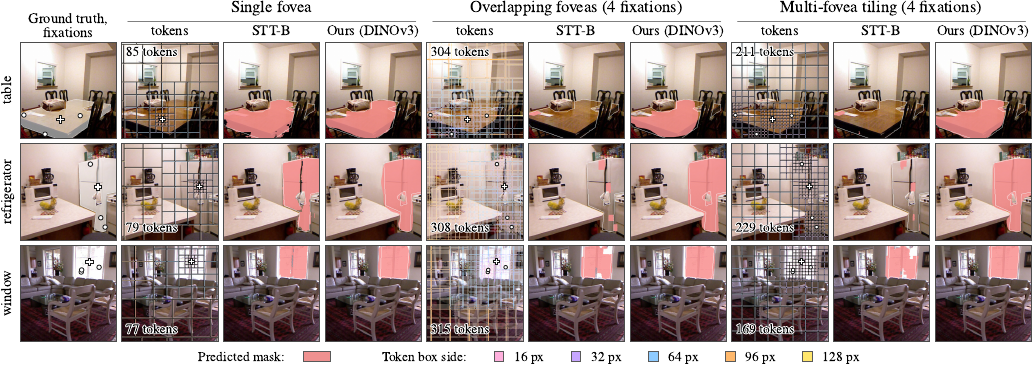}
    \caption{
    Comparison between \name{}-B and STT-B for single object segmentation \cite{schmidt2025stt} on NYUv2 \cite{nyudataset} segmentation dataset. Our method performs comparably for single fovea segmentation, and also generalises to multi-fovea settings. Neither model is trained on this dataset.}
    \label{fig:multi-fovea-qual}
    
\end{figure*}

\begin{figure*}[!t]
    \centering
    \includegraphics[width=0.95\linewidth]{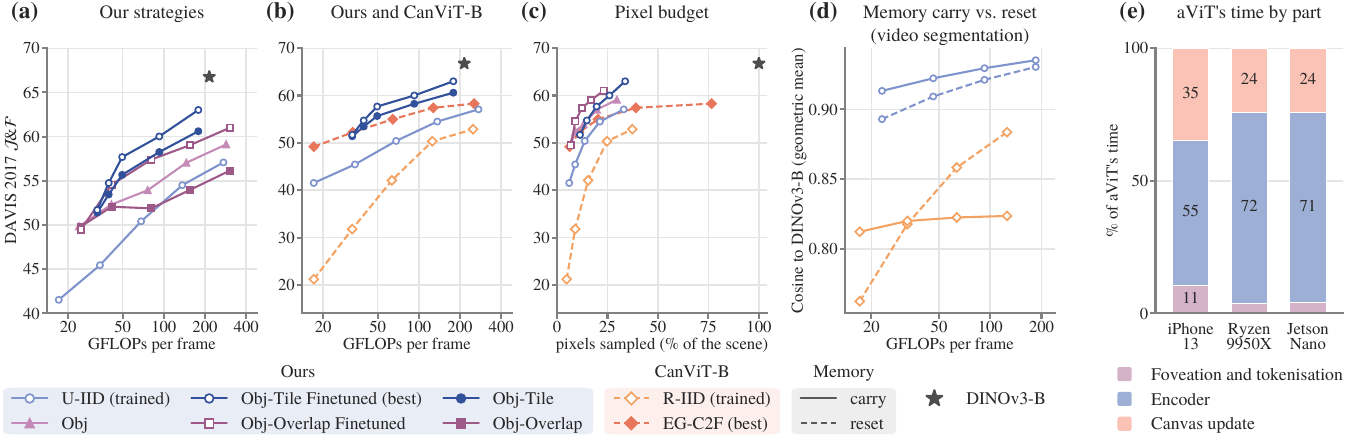}
    \caption{\textbf{Glimpse strategy, pixel budget, memory and latency.}
    Points on each line show 1, 2, 4, 8 and 16 glimpses, from left to right.
    (\textbf{a}) \name{}-B (DINOv3) with different foveation strategies.
    (\textbf{b}) Trained and best strategies of an \name{} and CanViT~\cite{berreby2026canvit}, with CanViT-B's memory reset at every frame. We include the YOLO computation in our measurements.
    (\textbf{c}) Performance against pixel budget, the fraction of the image observed~\cite{blauch2026a, pardyl2024adaglimpse, candes2008introduction}.
    (\textbf{d})  \name{} and CanViT with memory retained or reset at every frame.
    (\textbf{e}) Timing breakdown of the \name{} with DINOv3-B. \cref{fig:teaser} gives the total inference time on each device. } 
    \label{fig:huge_guy}
\end{figure*}

\subsection{Memory analysis}\label{sec:memory}
\name{}s keep a memory that each glimpse updates and that is carried from one frame to the next (\cref{fig:method}). Within a frame, each glimpse adds to what the memory holds. \cref{fig:depth_improvement} shows this for depth: after a second glimpse, the error falls, most of all around the new fovea. Corresponding quantitative results are shown in the supplementary. Across frames, the memory must stay useful while the scene changes. \cref{fig:huge_guy}\,(\textbf{d}) compares each model with its memory carried and with it reset at every frame. CanViT significantly underperforms when its memory is carried over than when it is reset. 

\subsection{Multiple foveation}
\label{sec:multifovea_exp}
We evaluate whether each model can use several fixations in one pass  (\cref{fig:multi-fovea-qual}). For each object in the NYUv2 \cite{nyudataset} test set, at 512\,px, we take four fixations: the centre of mass of its ground truth mask and three further points on the object.
We compare three inputs: a single fovea at the ground-truth centre of mass, four overlapping foveas, and one tiling of the four fixations (\cref{sec:multifovea}). Neither model was trained on NYUv2 or with multiple fixations. From a single fovea, both STT-B and our \name{} segment the object. With overlapping foveas or the tiling, STT-B does not generalise, while the masks of the \name{} stay the same or improve. Two factors may explain this. STT's learned position embedding is tied to the 172 slots of its pattern (\cref{sec:rope}), so it has no embedding for tokens outside that pattern. Replacing the encoder's RoPE with a learned embedding may also disturb features that the encoder learned with RoPE.

\cref{fig:huge_guy}\,(\textbf{a}) compares the policies of \cref{sec:glimpse_policies}, and panel (\textbf{c}) plots performance against the fraction of pixels observed~\cite{pardyl2024adaglimpse}, which measures the pixel budget of \cite{blauch2026a}. For the same fraction of pixels, an \name{} with U-IID matches the best CanViT policy once enough glimpses are allowed, and the object-based policies outperform it. Obj-Overlap performs better than Obj-Tile for the same fraction of pixels (\textbf{c}), but costs more computation, because its overlapping foveas produce more tokens (\textbf{a}).

\subsection{Computation time analysis}\label{sec:compute_analysis}
We measured the per-glimpse execution time of an \name{} on a CPU and several edge devices \cref{fig:teaser}. Our model, trained with the DINOv3-B teacher, runs 7--10 times faster. For fair comparison across foveation patterns, we report the average of 5 inference runs using patterns randomly sampled with the U-IID strategy.
Total computation time for 2 and 4 glimpses were obtained by tokenizing the image using a tile pattern (\cref{sec:multifovea}) and performing a single forward pass, taking 14.4 and 29.4 ms on the iPhone 13, respectively, which is computationally cheaper than performing 1 glimpse two or four times. This trade-off allows the system to achieve more accurate outputs at almost no additional cost, if its operating environment is slightly tolerant of latency.
\section{Conclusion}
\label{sec:conclusion}

We presented Active Vision Transformers, a general architecture for ViT-based active vision. Many tasks in robotics and edge applications only need detailed features around a few points of interest, such as grasping,  segmentation,  object tracking, 3D reconstruction and pose estimation. \name{}s allow a system to exploit any prior on where to look, whether it comes from a policy, a user prompt or a tracker, and supports multiple fixations per frame. By spending computation only on these regions, an active ViT is up to 10$\times$ cheaper to run than its standard counterpart. The method outperforms other active vision foundation models on a range of general tasks, and the reduction in computation enables real-time use on CPUs and mobile devices.

\clearpage
\section{LLM Usage Statement}
Although the direction of our ideas was developed entirely through discussions among the authors and collaborators, the process of determining the details involved discussions with Claude Code. We received extensive assistance from Claude Code throughout the entire process of writing code, downloading and processing experimental data, and actually conducting experiments, and a substantial portion of our system was written and maintained using Claude Code. For the main text, it was used to polish the wording or to translate drafts written in other languages.
\section{Acknowledgements}
We would like to thank Tanner Schmidt, David Sella, Justine Giroux, Eric Dexheimer, Jae-Young Kang and Jaeseok Jeong, for constructive and insightful discussions. This research was funded by the EPSRC On-Sensor Computer Vision grant (EP/Y020499/1, EP/Y023048/1), and used the resources of the Isambard-AI National AI Research Resource (AIRR)~\cite{mcintosh2024isambard}.

{
    \small
    \bibliographystyle{ieeenat_fullname}
    \bibliography{main}
}

\clearpage
\maketitlesupplementary
\appendix

\section{Implementation and evaluation details}
\label{sec:supp_impl}

\subsection{Training details}

\begin{figure}[t]
  \centering
  \includegraphics[width=\linewidth]{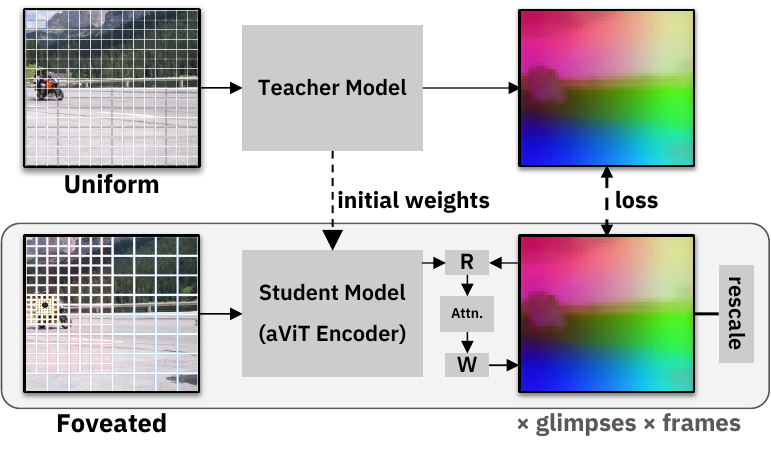}
  \caption{Teacher-student distillation}
  \label{fig:teacher-student-distillation}
\end{figure}

\begin{figure}[t]
    \centering
    \includegraphics[width=\linewidth]{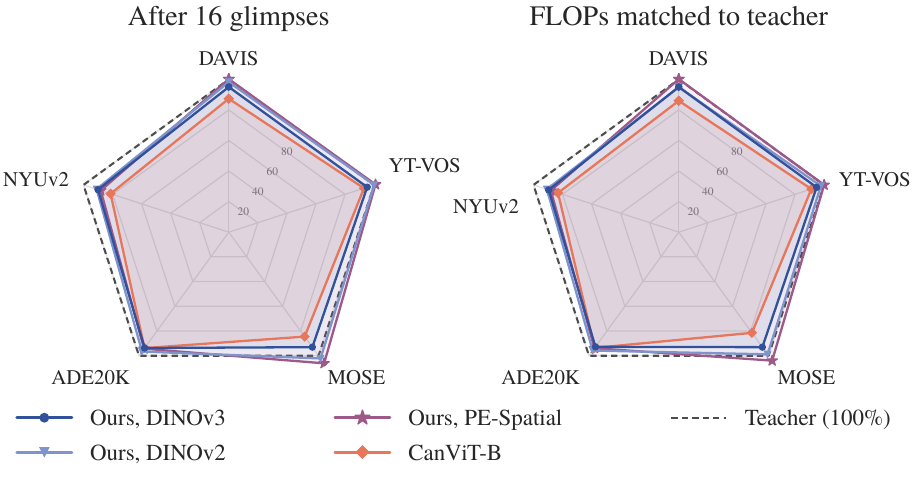}
    \caption{\textbf{Performance retention ratio} of our method with various models used as teachers compared against CanViT, both with \textit{memory being reset each frame}. For metrics where lower values are better, such as Depth RMSE (NYUv2 dataset), the reciprocal of this ratio is used.}
    \label{fig:pentagon_pair_reset}
\end{figure}

We initialise the encoder with the teacher's weights and fine-tune it with foveated tokens (\cref{fig:teacher-student-distillation}). The backbone therefore does not have to be trained from scratch, which is why different pre-trained ViTs can be converted with a short training phase (\cref{fig:pentagon_pair,fig:pentagon_pair_reset}).
The 512\,px \name{} of \cref{fig:huge_guy} is trained on SA-V for 100{,}000 steps at batch size 32, with 8 trained glimpses for each still and 6 for each clip. This amounts to 3.2 million sequences and about 22 million trained glimpses. Training takes about 22 hours on a single GH200, estimated as the step time multiplied by the number of steps.
CanViT-B distils the same teacher, but trains its backbone from a random initialisation. It uses 13.2 million ImageNet-21k~\cite{ridnik2021imagenetk} images and one billion glimpses, which its authors report takes 166 hours on a single H100~\cite{berreby2026canvit}. A GH200 combines an H100 GPU and a Grace CPU in one package, with shared, cache-coherent memory, so the two training times are roughly comparable. CanViT-B therefore trains about 7.5 times as long as the \name{}, on about 45 times as many glimpses.
The 20{,}000-step models of \cref{sec:ablations} and \cref{fig:imagenet_classification} are trained at batch size 8 and take about 1.5 hours on one RTX 5090.

\paragraph{Teachers.}
Besides DINOv3 ViT-B/16~\cite{simeoni2026dinov3}, we convert DINOv2-B~\cite{oquab2024dinov2}, PE-Spatial-B~\cite{bolya2026perception}, C-RADIOv3-B~\cite{ranzinger2024amradio}, iBOT-B~\cite{zhou2021ibot}, and DINOv3-L-originated encoder of VGGT-$\Omega$~\cite{wang2026vggtomega} (\cref{tab:fovit_teachers,fig:pentagon_pair,fig:pentagon_pair_reset,tab:vggt_omega_eth3d_7scenes}). We use C-RADIOv3-B rather than RADIOv2.5-B because it is the most recent RADIO model at this size and follows the RADIOv2.5 recipe. 

\subsection{Benchmark setting}\label{sec:benchmark_settings}

This section gives the settings for the tasks of the main paper: video object segmentation (DAVIS~\cite{pont2017davis, Perazzi_2016_CVPR}, YouTube-VOS~\cite{xu2018youtubevos, Xu_2018_ECCV} and MOSE~\cite{MOSE}), semantic segmentation (ADE20K~\cite{zhou2017scene}) and depth estimation (NYUv2~\cite{nyudataset}). Other tasks are described in \cref{sec:other_tasks}.
We follow the benchmark settings of CanViT for semantic segmentation on ADE20K, and those of DINOv3 for the other tasks, with the following differences.
Each image is cropped to a centre square, with its shorter side as the side length, and resized to $512\times 512$\,px.
The heads for semantic segmentation and depth estimation are trained on the canvas read at each glimpse of each model's training episodes, following the training rules of CanViT for ADE20K and of DINOv3 for NYUv2.
For each task, we choose the best policy by its metric on the training set.
In video object segmentation, the square crop can delay the frame at which an object first appears. In that case, the object's mask is given at the first frame in which it appears inside the square, and scoring starts from the next frame. This applies to 6.7\% of the videos in DAVIS, 12.6\% in YouTube-VOS and 41.5\% in MOSE.

\subsection{Measuring FLOPs}
  We report the cost of every model in GFLOPs, counted with PyTorch's
  \texttt{FlopCounterMode} during a forward pass on CPU, with one
  multiply-add counted as two FLOPs and the attention products included.
  All models are counted with the same code. CanViT~\cite{berreby2026canvit}
  uses the same convention and formulas. The cost of an \name{} glimpse depends
  on its position, because tokens that fall outside the image are dropped.
On a 512\,px frame, 77 of a glimpse's 172 tokens are inside the frame
    on average over 32 uniformly random positions, which costs 17.3 GFLOPs with the memory update included (16.8 GFLOPs averaged 
    over every position). Where the positions are not
  uniform we report the mean over the glimpses of the experiment; one glimpse
  at an object's centre of mass on NYUv2~\cite{nyudataset} uses 79 tokens on
  average and costs 17.8 GFLOPs. Task-specific heads are not included unless
  stated.

\section{Generality}
\label{sec:generality}

\subsection{Glimpse strategies}

\begin{figure}[t]
  \centering
  \includegraphics[width=\linewidth]{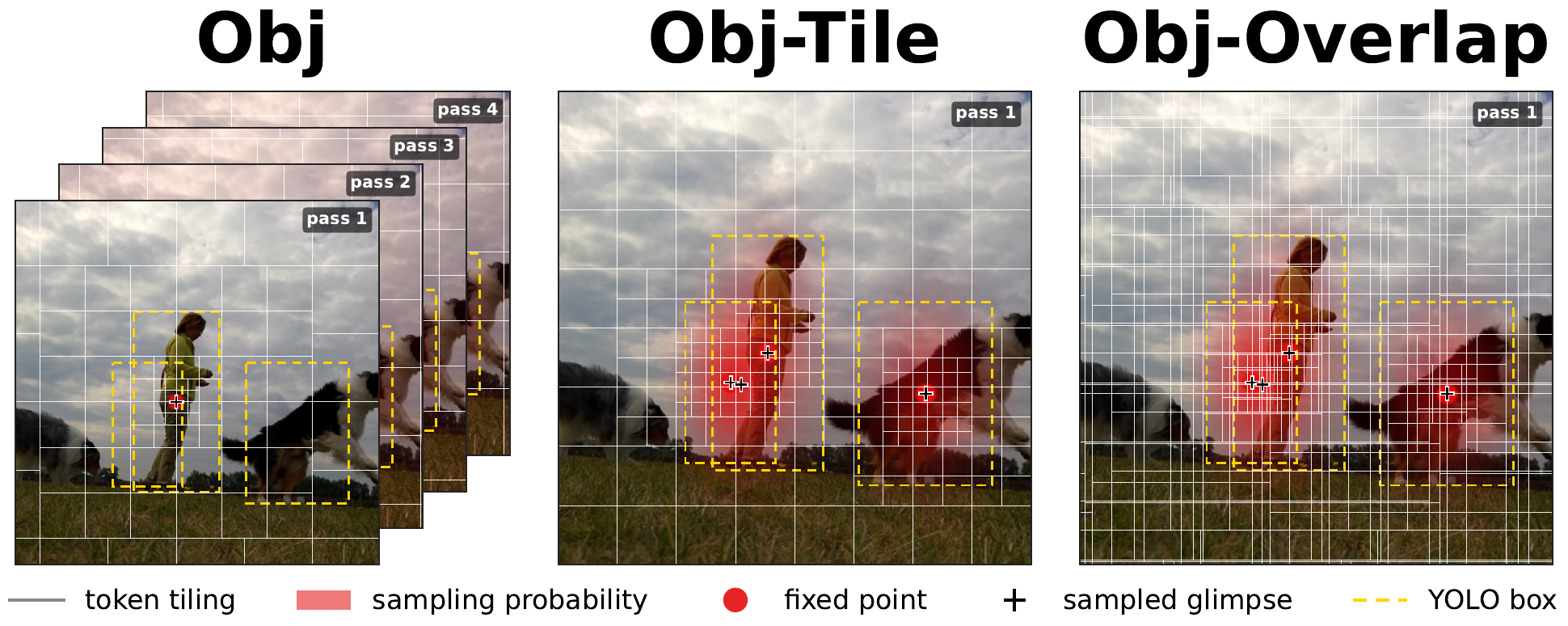}
  \caption{Fixation patterns and glimpse sampling distribution of object-centric sampling strategies.}
  \label{fig:multi-fovea-strategies}
  \vspace{-3mm}
\end{figure}

\cref{fig:multi-fovea-strategies} shows the object-based policies we use in our system. 
U-IID, F-IID and EG-C2F are described in the main paper.
The remaining CanViT policies are as follows.
R-IID draws windows of random size and position inside the image.
RFS repeats the view of the whole image.
C2F visits the tiles of a quadtree from coarse to fine, in random order within each level, and F2C reverses that sequence.
In the video tables, F2C reverses the $K$ glimpses of each frame, so with one glimpse it is the view of the whole image, as in C2F.
An \name{} glimpse has a fixation and no window, so for these policies we fixate at the centre of each window. RFS then repeats a fixation at the centre of the image.

We run Obj with both models. The released CanViT-B scores lower with Obj 
than with EG-C2F on every still image task and on every video benchmark with the memory reset
(\cref{fig:pentagon_pair_reset,tab:fovit_canvit}), and lower than with C2F on ImageNet (\cref{fig:imagenet_classification}).
Obj-Tile and Obj-Overlap combine several fixations in one glimpse, which CanViT-B cannot do, so we run them with an \name{} only.
Obj places the first glimpse of each frame at the centre of an object's bounding box, and moves to the next object at each frame.
Each later glimpse is drawn with probability one half from a normal distribution centred on an object's box, with a standard deviation of a quarter of the box's width and height. 
With the remaining half probability, it follows the U-IID distribution. Later glimpses of Obj-Overlap and Obj-Tile are always drawn from the box-centred Gaussian distribution.

\paragraph{Dependence on the strategy.} We train
\name{}s with U-IID only, yet they depend less on the test-time strategy than CanViT-B.
After one glimpse, its scores across strategies span 41.2 to 50.8 $\mathcal{J}\&\mathcal{F}$ on DAVIS and 38.7 to 42.9 mIoU on ADE20K, against 16.6 to 49.2 and 12.0 to 39.9 for CanViT-B (\cref{tab:fovit_strategies,tab:fovit_canvit}).
CanViT-B scores well only when its first glimpse covers the whole scene. An \name{} glimpse always includes a coarse periphery, so even a local first glimpse covers the scene.

\subsection{Other tasks}
\label{sec:other_tasks}

Beyond the tasks of the main paper, we evaluate \name{}s on correspondence finding (SPair-71k~\cite{min2019hyperpixel, min2019spair} and NAVI~\cite{jampani2023navi}), visual SLAM and structure from motion (ETH3D~\cite{schops2017multi,schops2019bad} and 7-Scenes~\cite{shotton2013scene}), and classification (ImageNet-1k~\cite{imagenet}). While not the focus of the paper, these experiments test the generality of \name{}s across tasks with minimal task-specific changes. Unless stated otherwise, we follow the settings of DINOv3, with the centre crop and resizing of \cref{sec:benchmark_settings}.

\paragraph{Correspondence finding.}
Cropping to a square removes some keypoints. On SPair-71k, we do not score a keypoint that lies outside the square in either image, which leaves 76.7\% of the keypoints. On NAVI, we do not apply the crop around the object, so some pairs have fewer object pixels than the number of matches scored per pair, and 99.3\% of the matches are scored. Results are in \cref{tab:fovit_teachers}.

\paragraph{Visual SLAM and SfM.}
We adapt the DINOv3-L encoder of VGGT-$\Omega$~\cite{wang2026vggtomega}, a feedforward 3D reconstruction model, into an \name{}-L using the same learning rate, step count and batch size as our other models (\cref{sec:supp_impl}).
We then replace the original encoder with the \name{}-L and fine-tune the aggregator with rank-16 LoRA.
Both stages use teacher-student distillation (\cref{fig:teacher-student-distillation}) on SA-V~\cite{ravi2024sam2}, DL3DV~\cite{ling2024dl3dv} and the egocentric videos of EgoExo4D~\cite{grauman2024ego}.
We evaluate the resulting pipeline on feedforward structure-from-motion using ETH3D~\cite{schops2017multi,schops2019bad} and 7-Scenes~\cite{shotton2013scene} (\cref{tab:vggt_omega_eth3d_7scenes}).
Applying KV-Tracker~\cite{Taher_2026_CVPR} to this pipeline produces the visual SLAM result in \cref{fig:teaser}.
This evaluation uses uniform random glimpses and separate training stages for the encoder and aggregator.
We leave task-specific glimpse policies and end-to-end fine-tuning to future work.

\begin{table*}[t]
\centering
\caption{Camera pose and depth with VGGT-$\Omega$ (1B, 512) on ETH3D and 7-Scenes, its own encoder against \name{}-L with the aggregator LoRA, from $K$ uniformly placed glimpses a frame (Sequential: one glimpse at a time; Tile: the $K$ foveae in one glimpse). Square $512\times512$ inputs (centre crop), 10 frames per scene (ETH3D, 13 scenes) or per test sequence (7-Scenes, 18 sequences), the official VGGT-$\Omega$ evaluation code. Bold: the best \name{}-L entry.}
\small
\setlength{\tabcolsep}{4pt}
\begin{tabular}{lcccccccccc}
\toprule
 & & & \multicolumn{4}{c}{ETH3D} & \multicolumn{4}{c}{7-Scenes} \\
\cmidrule(lr){4-7} \cmidrule(lr){8-11}
Encoder & Glimpses & $K$ & AUC@3$^\circ\uparrow$ & AUC@30$^\circ\uparrow$ & $\delta_{1.25}\uparrow$ & AbsRel$\downarrow$ & AUC@3$^\circ\uparrow$ & AUC@30$^\circ\uparrow$ & $\delta_{1.25}\uparrow$ & AbsRel$\downarrow$ \\
\midrule
VGGT-$\Omega$ & -- & -- & 54.0 & 92.0 & 99.9 & 0.008 & 27.2 & 83.6 & 97.7 & 0.026 \\
\midrule
\name{}-L & Sequential & 1 & 10.5 & 65.3 & 97.1 & 0.044 & 6.8 & 68.4 & 96.1 & 0.046 \\
\name{}-L & Sequential & 2 & 16.2 & 72.7 & 98.1 & 0.036 & 10.4 & 74.0 & 96.5 & 0.042 \\
\name{}-L & Sequential & 4 & 22.9 & 77.3 & 98.5 & 0.030 & 12.2 & 75.9 & 96.5 & 0.040 \\
\name{}-L & Sequential & 8 & \textbf{27.6} & \textbf{80.9} & \textbf{98.8} & \textbf{0.027} & \textbf{14.9} & \textbf{77.2} & 96.7 & 0.038 \\
\name{}-L & Tile & 2 & 20.6 & 76.9 & 98.4 & 0.030 & 10.8 & 74.0 & 96.7 & 0.039 \\
\name{}-L & Tile & 4 & 23.6 & 77.4 & \textbf{98.8} & \textbf{0.027} & 11.1 & 74.2 & \textbf{96.8} & \textbf{0.037} \\
\bottomrule
\end{tabular}
\label{tab:vggt_omega_eth3d_7scenes}
\end{table*}

\paragraph{Classification.}

\begin{figure*}
    \centering
    \includegraphics[width=1\linewidth]{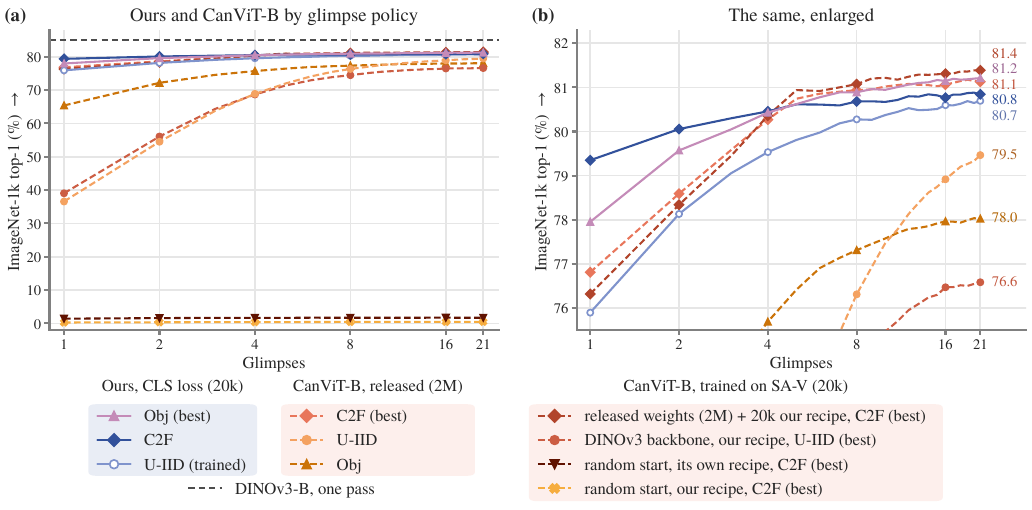}
    \caption{Performance of an \name{}-B vs CanViT-B on image classification. \textbf{Ours is trained 1.5 hours on a desktop GPU.} }
    \label{fig:imagenet_classification}
\end{figure*}

We also show \name{} performance on image classification (\cref{fig:imagenet_classification}).
We add a CLS output, and a small head that reads the encoder's CLS token at each glimpse and keeps a running average over glimpses.
It is trained to predict the teacher's CLS token for the whole frame, with the same CLS loss as CanViT~\cite{berreby2026canvit}, added to the patch loss with equal weight. 
Following CanViT's frozen protocol, we evaluate on 50{,}000 ImageNet-1k validation images~\cite{imagenet} at 512\,px by passing each model's predicted CLS token through the linear classifier of the DINOv3 teacher~\cite{simeoni2026dinov3}.
Neither model was trained on ImageNet-1k labels.
At 21 glimpses, the \name{}-B with its best policy (Obj) matches the released CanViT-B with its best policy (C2F), at 81.2\% and 81.1\% top-1 accuracy, against 84.9\% for the teacher on the whole image (\cref{fig:imagenet_classification}).
The \name{} does better for small numbers of glimpses, with 78.0\% against 76.8\% after one glimpse.
Our method also depends less on the policy: our three policies end within 0.5 points of each other, and CanViT-B's within 3.1 points.
We achieve this accuracy after 20{,}000 training iterations on SA-V~\cite{ravi2024sam2}, about 1.5 hours on one RTX 5090 (step time multiplied by the number of steps).
CanViT-B is trained for two million iterations on ImageNet-21k, which takes 166 hours on an H100~\cite{berreby2026canvit}.

\subsection{Multiple foveae}

\begin{figure}[t]
    \centering
    \includegraphics[width=0.9\linewidth]{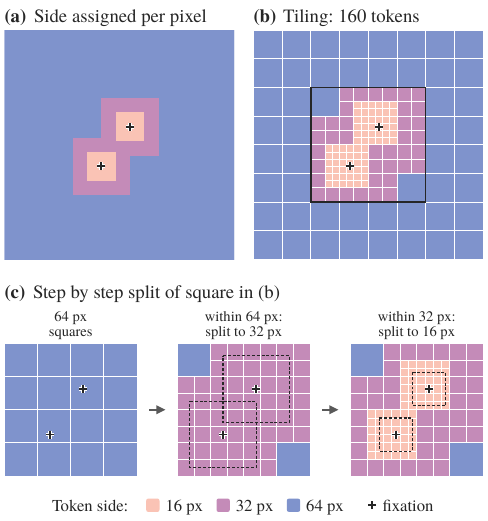}
        \caption{\textbf{Building a tiling for two fixations} (crosses) on a 512\,px frame.
    (\textbf{a})~The token side assigned to each pixel. 
    (\textbf{b})~The resulting tiling, with 160 tokens.
    (\textbf{c})~The outlined area of (b) after each split. Dashed lines mark the distance from a fixation within which a square is split. The steps start from 64\,px squares, because every 128\,px square of this frame is split.}
    \label{fig:tiling_construction}
    \vspace{-3mm}
\end{figure}

\paragraph{Constructing a tiling.}
\cref{tab:pattern} contains the foveation pattern of STT~\cite{schmidt2025stt}, which we use for a single fixation.
For several fixations, we adapt it into one tiling of the frame (\cref{sec:multifovea}), and \cref{fig:tiling_construction} shows the steps for two fixations.
First, each pixel is assigned a token side from its distance to the nearest fixation, measured as the larger of the horizontal and vertical offsets: 16\,px within 32\,px, 32\,px within 64\,px, 64\,px within 384\,px, and 128\,px beyond (\textbf{a}).
These are the distances of \cref{tab:pattern}, except that 64\,px tokens replace the 96\,px ones, because halving a 128\,px square can only give sides of 64, 32 and 16\,px.
Second, the frame is covered by $128 \times 128$\,px squares, and each square is split into quarters, repeatedly, for as long as any of its pixels has been assigned a smaller side (\textbf{c}).
The result covers every pixel with exactly one token (\textbf{b}).
In this example, the tiling for either fixation alone has 118 tokens and the tiling for both has 160, because the fixations share their coarse tokens.
The two STT patterns overlapped would have 179.
For a single fixation, the tiling has more tokens than the STT pattern (173 against 104 at 720\,px, \cref{tab:multifovea_cost}), for two reasons: the 96\,px ring uses smaller tokens, and the fine regions are rounded out to whole squares.

\begin{figure}[t]
  \centering
  \includegraphics[width=\linewidth]{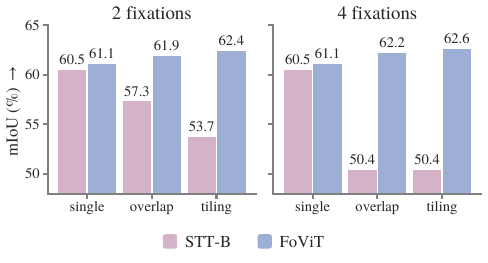}
  \caption{\textbf{Quantitative results for the experiment in \cref{fig:multi-fovea-qual}. Input tokens from several fixations, zero-shot.} Mean object-mask
  IoU (\%) over the 5{,}507 objects of the NYUv2 test split at 512\,px, for a
  single fovea, overlapping foveas and a tiling, with 2 fixations (left) and 4
  (right). All tokens of a set are given in one pass. Neither model was trained
  on such sets. STT-B loses up to 10 points when tokens from several fixations
  are combined, whereas our model gains slightly. The single-fixation bars are shown in both panels, STT and Ours achieve similar performance. Note that the y axis starts at 45 mIoU so differences appear larger.}
  \label{fig:multi-fovea-bars}
\end{figure}

\paragraph{Comparison with STT.} We show the details behind the multiple-fovea experiment of \cref{sec:multifovea_exp}. \cref{fig:multi-fovea-bars} shows the quantitative results for the experiment in \cref{fig:multi-fovea-qual}. With a single fovea, the two models perform similarly. When tokens from several fixations are combined, STT-B loses up to 10 points of IoU, while the \name{} gains slightly. Neither model was trained on such inputs for this experiment.

\cref{tab:multifovea_cost} evaluates the cost of additional fixations. The cost of overlapping foveas grows in proportion to their number (see Obj-Overlap, \cref{fig:multi-fovea-strategies}): four cost 77.3 GFLOPs in the encoder, about four times a single fovea. A tiling covers each region only once, so its cost grows more slowly.  Four tiled fixations cost 43.1 GFLOPs in the encoder and 54.1 GFLOPs with the memory update, which is 9.2 times cheaper than DINOv3 ViT-B on the full frame. \cref{tab:latency_glimpses} shows the latency for each of the multi-fixation patterns. Each one is the median of measurements made with 300 glimpse locations sampled randomly (uniform) in the image plane.

\begin{table}[t]
  \centering
  \caption{The 172 foveated tokens from ~\cite{schmidt2025stt}. Each is a $16 \times 16$ patch averaged over its box, and $\ell = \log_2(s/16)$ is its scale.}
  \small
  \begin{tabular}{@{}cccc@{}}
    \toprule
    Box side $s$ & Scale $\ell$ & Tokens & Distance from fixation \\
    \midrule
    16\,px  & 0    & 16 & 0--32\,px    \\
    32\,px  & 1    & 12 & 32--64\,px   \\
    64\,px  & 2    & 32 & 64--192\,px  \\
    96\,px  & 2.58 & 48 & 192--384\,px \\
    128\,px & 3    & 64 & 384--640\,px \\
    \bottomrule
  \end{tabular}
  \label{tab:pattern}
  \vspace{-7pt}
\end{table}

\begin{table*}[t]
  \centering
  \caption{Cost of one glimpse with several foveas, for an \name{} on a 720\,px frame: the encoder alone, and the encoder with the update of the canvas memory (one pass).\textbf{ Each cost is compared to the reduction relative to DINOv3 ViT-B on the same frame (499 GFLOPs)}. Tokens and GFLOPs are means over 5{,}507 NYUv2 objects, with all fixations on the object.}
  \label{tab:multifovea_cost}
  \begin{tabular}{lcrrrrr}
    \toprule
    & & & \multicolumn{2}{c}{Encoder} & \multicolumn{2}{c}{With memory} \\
    \cmidrule(lr){4-5} \cmidrule(lr){6-7}
    Glimpse & Foveas & Tokens & GFLOPs & Reduction & GFLOPs & Reduction \\
    \midrule
    Single fovea (baseline) & 1 & 104 & 19.0 & $26\times$  & 23.8 & $21\times$  \\
    \midrule
    Overlapping foveas      & 2 & 206 & 37.7 & $13\times$  & 47.3 & $11\times$  \\
                            & 4 & 410 & 77.3 & $6.5\times$ & 97.0 & $5.1\times$ \\
    \midrule
    Tiling                  & 1 & 173 & 31.7 & $16\times$  & 39.8 & $13\times$  \\
                            & 2 & 199 & 36.5 & $14\times$  & 45.8 & $11\times$  \\
                            & 4 & 234 & 43.1 & $12\times$  & 54.1 & $9.2\times$ \\
    \bottomrule
  \end{tabular}
\end{table*}

\paragraph{Fine-tuning for multiple fixations.}

\cref{tab:multifovea_iou} repeats the experiment of \cref{sec:multifovea_exp} on a 720\,px image. Without fine-tuning, a multi-tiling of the image improves slightly on the single fovea, and the gain grows with the number of fixations. This makes sense as the model is simply given higher resolution information around the target. Overlapping tilings give little or no gain, since the model is never taught to use the overlap. The model is trained with glimpses in sequence: four sequential glimpses therefore do better than tiling and overlapping.  

We then fine-tuned the model for a further 15{,}000 steps on glimpses with one to four fixations.  After fine-tuning, both kinds of combined glimpse improve with the number of fixations. The tiling of four fixations then slightly exceeds four glimpses in sequence, with about 60\% of the tokens and compute. Fine-tuning does not lower the IoU of the single fovea or of glimpses in sequence. This shows that our architecture adapts to different tile sizes. Future work could explore post-training for non-square or other adaptive tokenisations.

\begin{table*}[t]
  \centering
  \caption{Mask IoU on 5{,}507 NYUv2 objects for our model trained at 720\,px, before and after fine-tuning on glimpses with several
  foveas (720\,px frame, one pass a glimpse). The first fixation is at the centre of mass of the object's ground-truth mask and the
  others are random pixels of the mask. In the first three groups all foveas are encoded in one glimpse; the last row is four
  glimpses in sequence with one fovea each. Tokens is the mean over the objects (for the last row, the total over the four glimpses).
  $\Delta$ is the change from the same model's single fovea, computed before rounding; its standard error is about 0.001. The
  fine-tuned model was trained for 15{,}000 further steps on such token sets.}
  \label{tab:multifovea_iou}
  \begin{tabular}{lcrrrrr}
    \toprule
    & & & \multicolumn{2}{c}{Final model} & \multicolumn{2}{c}{Fine-tuned} \\
    \cmidrule(lr){4-5} \cmidrule(lr){6-7}
    Glimpse & Foveas & Tokens & IoU & $\Delta$ & IoU & $\Delta$ \\
    \midrule
    Single fovea (baseline) & 1 & 104 & 0.648 &          & 0.650 &          \\
    \midrule
    Overlapping foveas      & 2 & 206 & 0.653 & $+0.005$ & 0.664 & $+0.014$ \\
                            & 4 & 410 & 0.648 & $0.000$  & 0.671 & $+0.021$ \\
    \midrule
    Tiling                  & 1 & 173 & 0.656 & $+0.008$ & 0.665 & $+0.015$ \\
                            & 2 & 199 & 0.658 & $+0.010$ & 0.670 & $+0.020$ \\
                            & 4 & 234 & 0.661 & $+0.013$ & 0.676 & $+0.026$ \\
    \midrule
    Single glimpses in sequence & 4 & 410 & 0.671 & $+0.023$ & 0.672 & $+0.023$ \\
    \bottomrule
  \end{tabular}
\end{table*}

\section{Further analysis}
\label{sec:analysis}

\subsection{Computational analysis}
We measured the latency of an \name{}-B and CanViT-B for a single glimpse on a 9950X CPU in fp32. CanViT-B takes 48.8\,ms, about 20\% more than the equivalent \name{}. This difference is due to CanViT-B's readout stage, which projects its higher-dimensional intermediate features into the DINOv3 feature space and takes roughly one third of its latency. \name{}s do not need this readout stage, because its memory keeps track of features directly in the teacher feature space, up to normalisation.

\cref{tab:latency} compares the latency of an \name{} and the teacher model on various devices. \name{}s have a particular advantage over the teacher model on processors with limited parallelism, such as edge processors or CPUs (7--10$\times$ faster). On desktop GPUs, the gain is smaller (2.5--3.5$\times$ faster); this is because the attention between all tokens, the main bottleneck that \name{}s resolve by reducing the number of tokens, can already be computed sufficiently fast on more massively parallel processors such as desktop GPUs. Reported values are medians over 300 glimpse locations sampled uniformly at random in the image plane.

\begin{table}[t]
  \centering
  \caption{\textbf{Runtime of each stage in ms (median)} for one glimpse and one pass.}
  \small
  \sisetup{mode=text}
  \setlength{\tabcolsep}{3.5pt}
  \begin{adjustbox}{max width=\linewidth}
    \begin{tabular}{@{}l
                    S[table-format=2.2]
                    S[table-format=4.2]
                    S[table-format=2.2]
                    S[table-format=4.2]@{}}
      \toprule
            & {Foveation \&} &           &          &         \\
      Model & {Tokenisation} & {Encoder} & {Memory} & {Total} \\
      \midrule
      \multicolumn{5}{@{}l}{\textit{Apple A15 Bionic (Neural Engine, fp16)}} \\
      \name{}-L            &  0.99 &   20.8 &  4.9 &   26.7 \\
      Teacher (\cite{wang2026vggtomega}'s ViT-L)    &  1.04 &  323.4 & {--} &  325.1 \\
      \name{}-B            &  1.15 &    5.9 &  3.8 &   10.9 \\
      Teacher (DINOv3-B) &  0.94 &   92.8 & {--} &   93.8 \\
      \midrule
      \multicolumn{5}{@{}l}{\textit{AMD Ryzen 9 9950X (CPU, fp32)}} \\
      \name{}-L            &  2.00 &   93.6 & 14.4 &  110.5 \\
      Teacher (\cite{wang2026vggtomega}'s ViT-L)    &  1.92 &  943.2 & {--} &  945.3 \\
      \name{}-B            &  1.51 &   28.1 &  9.5 &   39.2 \\
      Teacher (DINOv3-B) &  1.66 &  282.0 & {--} &  283.7 \\
      \midrule
      \multicolumn{5}{@{}l}{\textit{NVIDIA Tegra X1 (Maxwell GPU, fp32)}} \\
      \name{}-L            &  7.48 &  405.5 & 61.6 &  474.5 \\
      Teacher (\cite{wang2026vggtomega}'s ViT-L)    & 11.12 & 5044.4 & {--} & 5055.5 \\
      \name{}-B            &  7.13 &  118.7 & 40.2 &  166.0 \\
      Teacher (DINOv3-B) &  8.68 & 1620.4 & {--} & 1629.1 \\
      \midrule
      \multicolumn{5}{@{}l}{\textit{NVIDIA GeForce RTX 5090 (Blackwell GPU, fp32)}} \\
      \name{}-L            &  0.53 &   4.71 & 0.66 &   5.87 \\
      Teacher (\cite{wang2026vggtomega}'s ViT-L)    &  0.09 &  22.44 & {--} &  22.48 \\
      \name{}-B            &  0.51 &   2.35 & 0.58 &   3.41 \\
      Teacher (DINOv3-B) &  0.07 &   9.12 & {--} &   9.18 \\
      \bottomrule
    \end{tabular}
  \end{adjustbox}
  \label{tab:latency}
\end{table}

\begin{table*}[t]
  \centering
  \caption{\textbf{Total runtime in ms (median) of one frame, one pass}: a standard single glimpse, and 1, 2 and 4 fixations encoded together as Obj-Overlap (each glimpse's own pattern) and Obj-Tile (one tiling of the frame for all glimpses).}
  \small
  \sisetup{mode=text}
  \begin{tabular*}{\textwidth}{@{\extracolsep{\fill}}l
                  S[table-format=3.2]
                  S[table-format=3.2]
                  S[table-format=4.2]
                  S[table-format=4.2]
                  S[table-format=3.2]
                  S[table-format=3.2]
                  S[table-format=4.2]@{}}
    \toprule
                     & \text{Single fovea} & \multicolumn{3}{c}{Obj-Overlap} & \multicolumn{3}{c}{Obj-Tile} \\
    \cmidrule(lr){3-5} \cmidrule(l){6-8}
    Model / glimpses & {1} & {1} & {2} & {4} & {1} & {2} & {4} \\
    \midrule
    \multicolumn{8}{@{}l}{\textit{Apple A15 Bionic (Neural Engine, fp16)}} \\
    \name{}-L &   26.6 &   26.6 &   34.1 &   73.8 &   27.7 &   32.2 &   55.5 \\
    \name{}-B   &   10.9 &   10.9 &   14.5 &   29.3 &   11.7 &   14.0 &   20.5 \\
    \midrule
    \multicolumn{8}{@{}l}{\textit{AMD Ryzen 9 9950X (CPU, fp32)}} \\
    \name{}-L &  113.1 &  113.1 &  181.3 &  322.9 &  141.0 &  166.0 &  204.3 \\
    \name{}-B   &   40.6 &   40.6 &   63.0 &  110.4 &   50.4 &   56.9 &   70.2 \\
    \midrule
    \multicolumn{8}{@{}l}{\textit{NVIDIA Tegra X1 (Maxwell GPU, fp32)}} \\
    \name{}-L &  474.5 &  474.5 &  957.2 & 1582.7 &  484.5 &  939.2 & 1474.4 \\
    \name{}-B   &  166.0 &  166.0 &  329.6 &  558.5 &  167.5 &  315.8 &  499.0 \\
    \midrule
    \multicolumn{8}{@{}l}{\textit{NVIDIA GeForce RTX 5090 (Blackwell GPU, fp32)}} \\
    \name{}-L &   6.13 &   6.13 &   9.58 &  13.26 &   8.30 &   8.64 &  10.66 \\
    \name{}-B   &   3.39 &   3.39 &   4.67 &   6.17 &   5.19 &   4.70 &   5.03 \\
    \bottomrule
  \end{tabular*}
  \label{tab:latency_glimpses}
\end{table*}

\subsection{Memory}

\begin{figure*}[t]
  \centering
  \includegraphics[width=0.9\linewidth]{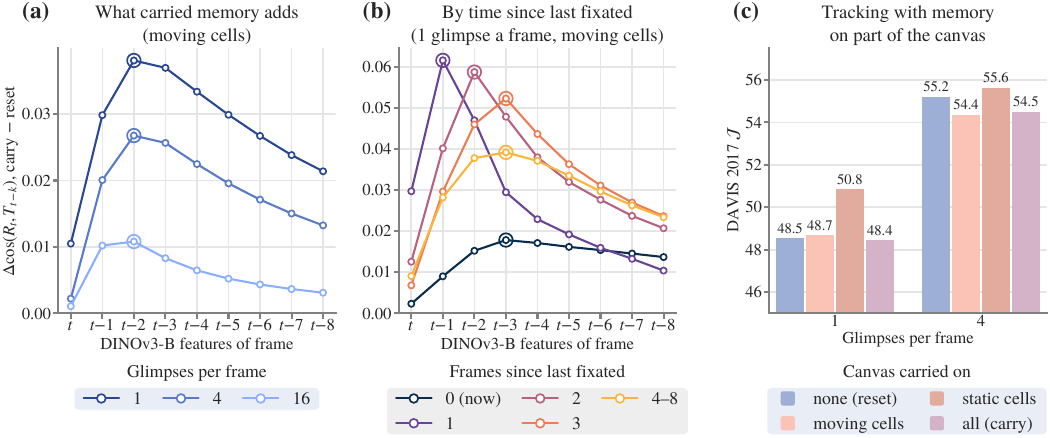}
\caption{Memory lag analysis. The memory settles to the features of a few frames earlier.}
  \label{fig:memory_lag}
\end{figure*}

\paragraph{Temporal lag in dynamic scenes.}
The canvas is carried from frame to frame.
\cref{fig:memory_lag} shows that this helps in slow-moving regions, but the canvas lags fast-moving objects by 1 to 3 frames. Motion compensation on the canvas and learned gating could improve the accuracy, which we leave for future work.
    \vspace{-4mm}

\begin{figure}[t]
  \centering
  \includegraphics[width=\linewidth]{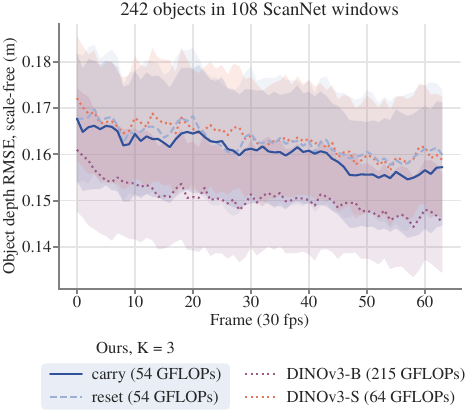}
  \caption{Depth RMSE measured on object mask of ScanNetv2 \cite{dai2017scannet}.}
  \label{fig:depth_per_object}
\end{figure}

\paragraph{Benefits of memory.}
While carried memory lags on fast-moving objects, it helps in slowly moving scenes.
On ScanNetv2~\cite{dai2017scannet}, we select video segments of at least 60 frames (2\,s at 30\,fps) where the camera pose changes slowly.
We apply the Obj strategy using the tightest axis-aligned bounding box around each ground-truth segmentation mask and measure the object depth RMSE.
\cref{fig:depth_per_object} shows that carrying the canvas achieves lower depth RMSE than resetting it at every frame, whereas resetting matches DINOv3-S.
Under slow camera motion, propagating the canvas across frames retains useful features from earlier views and outperforms DINOv3-S, a smaller distilled version of DINOv3-B.

\section{Ablation}
\label{sec:ablations}

\begin{table*}[t]
\centering
\caption{\textbf{Ablations on DAVIS 2017 val.} $\mathcal{J}\&\mathcal{F}$ (higher is better) with the canvas carried from frame to frame, for 1, 2, 4, 8 and 16 glimpses a frame at uniformly random positions (U-IID), and the mean over the five. Each model differs from the baseline only as 
stated. \textbf{Every model here, the baseline included, is trained for only 20k iterations at batch size 8}; the final model of 
the main paper is trained for 100k iterations at batch size 32. \textbf{GFLOPs are for one glimpse}. Differences below 0.7 at a single number of glimpses may be due to the seed alone (\cref{sec:ablations}). Bold: best in each column. Underlined: second best.}
\label{tab:ablations-carried}
\begin{tabular}{@{}l r rrrrrr@{}}
\toprule
 & & \multicolumn{6}{c}{$\mathcal{J}\&\mathcal{F}$ by glimpses a frame, canvas carried $\uparrow$} \\
\cmidrule(l){3-8}
Model & GFLOPs & 1 & 2 & 4 & 8 & 16 & Mean \\
\midrule
Baseline (RoPE scale 8) & 17.3 & 35.8 & 39.7 & 43.4 & 48.4 & \textbf{53.6} & \textbf{44.2} \\
\midrule
\multicolumn{8}{@{}l}{\textit{RoPE scale (baseline 8)}} \\
\quad Scale 1 & 17.3 & 35.0 & 38.7 & 41.5 & 45.0 & 49.5 & 41.9 \\
\quad Scale 2 & 17.3 & 34.8 & 39.0 & 42.9 & 48.3 & 52.8 & \underline{43.6} \\
\quad Scale 4 & 17.3 & 35.2 & 39.6 & 43.9 & \textbf{48.8} & \underline{53.5} & \textbf{44.2} \\
\quad Scale 16 & 17.3 & \textbf{36.6} & 40.0 & 43.7 & 48.0 & 52.9 & \textbf{44.2} \\
\quad Scale 32 & 17.3 & 35.5 & 39.7 & 43.1 & 47.4 & 50.8 & 43.3 \\
\midrule
\multicolumn{8}{@{}l}{\textit{Canvas rescaling and damage}} \\
\quad Single rescaling factor, not per cell & 17.3 & 35.7 & \underline{40.2} & 43.8 & \underline{48.5} & 52.7 & \textbf{44.2} \\
\quad No rescaling & 17.3 & 25.3 & 26.7 & 27.2 & 28.3 & 29.5 & 27.4 \\
\quad No damage in training & 17.3 & \underline{35.9} & 40.0 & 42.2 & 46.5 & 52.5 & 43.4 \\
\midrule
\multicolumn{8}{@{}l}{\textit{Read, write and self-attention block}} \\
\quad No self-attention block & 15.1 & 31.5 & 34.3 & 37.0 & 39.4 & 41.4 & 36.7 \\
\quad No self-attention block, w null slot & 15.1 & 31.3 & 33.3 & 34.5 & 34.5 & 33.4 & 33.4 \\
\quad Write only (no read, no block) & 14.7 & 31.3 & 37.2 & 42.5 & 43.7 & 44.1 & 39.8 \\
\quad Standard cross-attn.   & 26.9 & 35.0 & \textbf{40.4} & \textbf{46.0} & 47.7 & 48.9 & \underline{43.6} \\
\quad Standard cross-attn, w/o MLP & 17.1 & 31.3 & 38.4 & \underline{44.2} & 46.6 & 47.2 & 41.5 \\
\bottomrule
\end{tabular}
\end{table*}

\cref{tab:ablations-carried} ablates the main design choices of \name{}s. Every model is trained for 20k steps with batch size 8 and the same seed. A second seed of the baseline changes $\mathcal{J}\&\mathcal{F}$ only up to 0.7 at a single number of glimpses and by almost 0 in the mean, so differences smaller than 0.7 can be considered noise.

\subsection{RoPE scale}
\label{sec:abl_rope}

In the read and the write, RoPE positions are $(\mathit{row}, \mathit{col}) \in [-1, 1]^2$
over the image. In the self-attention block they are offsets from the fixation in the same units. We multiply them by 8 before
the rotation (see \cref{tab:ablations-carried}).

The shortest RoPE wavelength is then $1/16$ of the image, or two canvas cells, against $1/2$ of the image at scale 1, which CanViT~\cite{berreby2026canvit} uses.  Any scale from 4 to 16 works equally well. 
A possible reason, which we did not test directly, is that at scale 1 rotation angles of different canvas
cells become indistinguishable. RoPE rotates each pair of channels of a query or key by an angle proportional to its position along one axis, with a
different wavelength for each pair. The attention score depends on position only through the difference between the query's angle
and the key's angle. 
If a pair's wavelength is much longer than the image is wide, its angle is almost the same at every position in the image, so the pair gives almost the same score for every cell. We also tried applying RoPE based on $\ell$ (see \cref{tab:pattern}), but found the model learned relative token size without this.

\subsection{Canvas rescaling}
\label{sec:abl_rescaling}

After every write, we rescale each canvas cell to RMS 1. Otherwise the cell values grow with every write, and later writes change
  them less and less. Without rescaling, the mean $\mathcal{J}\&\mathcal{F}$ falls from 44.2 to 27.4 (see \cref{tab:ablations-carried}), and more glimpses add little.  Rescaling the whole canvas by a single factor per image gives the same mean as rescaling per cell, so on this benchmark the rescaling matters but its granularity does not.

\subsection{Canvas damage}
\label{sec:abl_damage}

Following Solv3R~\cite{kwak2026solv3r}, which perturbs its state with noise or a partial reset, we damage the canvas of each training sequence with a 15\% chance, before one of its glimpses. The damage takes one of three forms: random
noise, a reset of some cells to their initial values, or a shift of the whole canvas by up to 3 cells in each direction. Each form is similar to an error the canvas can meet at test time. Noise imitates accumulated errors, reset cells imitate regions that have not been seen, and a shift imitates camera motion that the canvas has not followed. The model therefore learns to correct the canvas from later glimpses rather than trust what it holds. Training without damage lowers the mean $\mathcal{J}\&\mathcal{F}$ by 0.8, with the largest drops at 4 and 8 glimpses (1.2 and 1.9).

\subsection{Read, write and self-attention block}
\label{sec:abl_write}

We tested different versions of the architecture before arriving at the design shown in \cref{fig:method}.  
Without layers of self-attention on the encoder output,  performance is much lower: $\mathcal{J}\&\mathcal{F}$ goes from 44.2 to 36.7 (\cref{tab:ablations-carried}) and the score no longer improves beyond 4 glimpses. We also try adding a null slot to the write, a learned key with a zero value, so that a cell can learn to take nothing from a glimpse, which actually degraded performance.

A write only has a linear map on the keys and values, which act on the glimpse tokens; the query is used directly (see \cite{berreby2026canvit}). We compare it with a standard cross-attention layer, in which the canvas cells attend to the glimpse tokens with learned query, key, value and output maps and a residual connection, followed by an MLP. This works better for 2 and 4 glimpses, but worse at 16, and it raises the cost of a glimpse from 17.3 to 26.9 GFLOPs. Without the MLP, the cost is close to the final method  (\cref{fig:method}), and the mean is only 2.7 lower. Future work could be to simplify the architecture and use only standard cross attention. 

\section{Experiment Tables}
In \cref{tab:fovit_canvit,tab:fovit_teachers,tab:fovit_strategies}, the numbers above the columns are the number of glimpses $K$; on still images, the canvas is read after $K$ glimpses. Obj-Tile and Obj-Overlap count fixations instead, up to four in one glimpse. The whole-scene values are those of the teacher, applied once to the whole 512\,px scene. Superscripts mark the finetuned \name{} (multi-fovea finetuning, 10k steps): \textsuperscript{T} (Tile) is finetuned with tiled fixations and \textsuperscript{O} (Overlap) with overlapping fixations (Obj is its one-fixation glimpse). In each column, the best value is in bold and the second best is underlined.

\begin{table}[t]
  \centering
  \setlength{\abovecaptionskip}{0pt}
  \caption{\name{}-B and CanViT-B with every strategy, DINOv3-B as the teacher. Video: $K$ glimpses a frame, the memory carried across frames. CanViT-B takes one fixation a glimpse, so it has no Obj-Tile or Obj-Overlap (--).}
  \label{tab:fovit_strategies}
  \vspace{4pt}
  \footnotesize
  \setlength{\tabcolsep}{2.5pt}
  \begin{adjustbox}{max width=\linewidth}
    \begin{tabular}{@{}l@{\hspace{9pt}}*{3}{w{c}{1.85em}}@{\hspace{9pt}}*{3}{w{c}{1.85em}}@{\hspace{9pt}}*{3}{w{c}{1.85em}}@{\hspace{9pt}}*{3}{w{c}{1.85em}}@{}}
      \toprule
       & \multicolumn{6}{@{}c@{\hspace{9pt}}}{DAVIS ($\mathcal{J}\&\mathcal{F}$ $\uparrow$), memory carry} & \multicolumn{6}{@{}c@{}}{ADE20K (mIoU $\uparrow$)} \\
      \cmidrule(r{9pt}){2-7} \cmidrule{8-13}
       & \multicolumn{3}{@{}c@{\hspace{9pt}}}{\name{}-B} & \multicolumn{3}{@{}c@{\hspace{9pt}}}{CanViT-B} & \multicolumn{3}{@{}c@{\hspace{9pt}}}{\name{}-B} & \multicolumn{3}{@{}c@{}}{CanViT-B} \\
      \cmidrule(r{9pt}){2-4} \cmidrule(r{9pt}){5-7} \cmidrule(r{9pt}){8-10} \cmidrule{11-13}
      Strategy & 1 & 4 & 16 & 1 & 4 & 16 & 1 & 4 & 16 & 1 & 4 & 16 \\
      \midrule
      \textit{whole scene} & \multicolumn{6}{@{}c@{\hspace{9pt}}}{66.8} & \multicolumn{6}{@{}c@{}}{48.6} \\
      \midrule
      Obj & 47.0 & 52.4 & 59.7 & \underline{19.5} & 20.3 & \underline{20.9} & 40.3 & 43.6 & \textbf{46.1} & 17.3 & 31.5 & 39.8 \\
      U-IID & 42.3 & 48.7 & 57.8 & 18.2 & 18.7 & 20.3 & 38.7 & 43.0 & 45.6 & 14.0 & 29.9 & 39.7 \\
      \cmidrule{1-13}
      F-IID & 41.2 & 51.0 & 58.3 & \textbf{20.5} & \underline{21.0} & \textbf{21.2} & 40.7 & 43.7 & 45.8 & \textbf{39.9} & 43.1 & 44.5 \\
      R-IID & 44.9 & 52.2 & 57.9 & 19.2 & 20.5 & 19.9 & 40.5 & 44.3 & \underline{46.0} & \underline{19.3} & 34.9 & 42.4 \\
      RFS & 41.2 & 41.9 & 42.9 & \textbf{20.5} & 20.6 & 20.5 & 40.7 & 40.9 & 40.8 & \textbf{39.9} & 40.9 & 39.7 \\
      F2C & 41.2 & 46.6 & 56.9 & \textbf{20.5} & 20.7 & 20.2 & 38.3 & 43.0 & 45.9 & 12.0 & 26.4 & 39.5 \\
      C2F & 41.2 & 47.1 & 57.3 & \textbf{20.5} & \underline{21.0} & \underline{20.9} & 40.7 & 44.2 & 45.8 & \textbf{39.9} & \underline{44.1} & \underline{45.4} \\
      EG-C2F & 41.2 & 46.7 & 57.0 & \textbf{20.5} & \textbf{21.1} & \underline{20.9} & 40.7 & \underline{44.4} & \textbf{46.1} & \textbf{39.9} & \textbf{44.3} & \textbf{45.6} \\
      \cmidrule{1-13}
      Obj-Tile & \underline{49.9} & 53.7 & \underline{60.9} & \multicolumn{3}{@{}c@{\hspace{9pt}}}{--} & \underline{42.6} & 43.4 & 44.9 & \multicolumn{3}{@{}c@{}}{--} \\
      Obj-Overlap & 47.0 & 48.1 & 54.2 & \multicolumn{3}{@{}c@{\hspace{9pt}}}{--} & 40.3 & 39.5 & 42.1 & \multicolumn{3}{@{}c@{}}{--} \\
      \midrule
      \multicolumn{13}{@{}l@{}}{\textit{\name{} (finetuned)}} \\
      Obj\textsuperscript{O} & 46.6 & 51.8 & 58.7 & \multicolumn{3}{@{}c@{\hspace{9pt}}}{--} & 39.9 & 43.4 & 45.9 & \multicolumn{3}{@{}c@{}}{--} \\
      Obj-Tile\textsuperscript{T} & \textbf{50.8} & \textbf{55.4} & \textbf{61.6} & \multicolumn{3}{@{}c@{\hspace{9pt}}}{--} & \textbf{42.9} & \textbf{44.8} & \underline{46.0} & \multicolumn{3}{@{}c@{}}{--} \\
      Obj-Overlap\textsuperscript{O} & 46.6 & \underline{54.6} & 60.5 & \multicolumn{3}{@{}c@{\hspace{9pt}}}{--} & 39.8 & 44.0 & 45.8 & \multicolumn{3}{@{}c@{}}{--} \\
      \bottomrule
    \end{tabular}
  \end{adjustbox}
\end{table}

\begin{table*}[t]
  \centering
  \setlength{\abovecaptionskip}{0pt}
  \caption{An \name{}-B  with DINOv3-B as the teacher and CanViT-B. Video: $K$ glimpses a frame, the memory reset every frame. Strategies: video segmentation, semantic segmentation, depth estimation and correspondence: Obj-Tile\textsuperscript{T} for \name{}-B and EG-C2F for CanViT. CanViT (our training): trained on SA-V by our protocol (150k steps) from the released weights or from DINOv3 ViT-B/16.}
  \label{tab:fovit_canvit}
  \vspace{4pt}
  \footnotesize
  \setlength{\tabcolsep}{2.5pt}
  \begin{adjustbox}{max width=\linewidth}
    \begin{tabular}{@{}l@{\hspace{9pt}}*{3}{w{c}{1.92em}}@{\hspace{9pt}}*{3}{w{c}{2.14em}}@{\hspace{9pt}}*{3}{w{c}{1.87em}}@{\hspace{9pt}}*{3}{w{c}{2.21em}}@{\hspace{9pt}}*{3}{w{c}{2.35em}}@{\hspace{9pt}}*{3}{w{c}{2.15em}}@{\hspace{9pt}}*{3}{w{c}{1.85em}}@{}}
      \toprule
       & \multicolumn{9}{@{}c@{\hspace{9pt}}}{Video seg., memory reset} & \multicolumn{3}{@{}c@{\hspace{9pt}}}{Sem. seg.} & \multicolumn{3}{@{}c@{\hspace{9pt}}}{Depth est.} & \multicolumn{6}{@{}c@{}}{Correspondence} \\
      \cmidrule(r{9pt}){2-10} \cmidrule(r{9pt}){11-13} \cmidrule(r{9pt}){14-16} \cmidrule{17-22}
       & \multicolumn{3}{@{}c@{\hspace{9pt}}}{DAVIS ($\mathcal{J}\&\mathcal{F}$ $\uparrow$)} & \multicolumn{3}{@{}c@{\hspace{9pt}}}{YT-VOS ($\mathcal{J}\&\mathcal{F}$ $\uparrow$)} & \multicolumn{3}{@{}c@{\hspace{9pt}}}{MOSE ($\mathcal{J}\&\mathcal{F}$ $\uparrow$)} & \multicolumn{3}{@{}c@{\hspace{9pt}}}{ADE20K (mIoU $\uparrow$)} & \multicolumn{3}{@{}c@{\hspace{9pt}}}{NYUv2 (RMSE $\downarrow$)} & \multicolumn{3}{@{}c@{\hspace{9pt}}}{SPair-71k (PCK $\uparrow$)} & \multicolumn{3}{@{}c@{}}{NAVI (Recall $\uparrow$)} \\
      \cmidrule(r{9pt}){2-4} \cmidrule(r{9pt}){5-7} \cmidrule(r{9pt}){8-10} \cmidrule(r{9pt}){11-13} \cmidrule(r{9pt}){14-16} \cmidrule(r{9pt}){17-19} \cmidrule{20-22}
      Model & 1 & 4 & 16 & 1 & 4 & 16 & 1 & 4 & 16 & 1 & 4 & 16 & 1 & 4 & 16 & 1 & 4 & 16 & 1 & 4 & 16 \\
      \midrule
      DINOv3-B (\textit{whole scene}) & \multicolumn{3}{@{}c@{\hspace{9pt}}}{66.8} & \multicolumn{3}{@{}c@{\hspace{9pt}}}{65.8} & \multicolumn{3}{@{}c@{\hspace{9pt}}}{40.7} & \multicolumn{3}{@{}c@{\hspace{9pt}}}{48.6} & \multicolumn{3}{@{}c@{\hspace{9pt}}}{0.400} & \multicolumn{3}{@{}c@{\hspace{9pt}}}{54.1} & \multicolumn{3}{@{}c@{}}{57.0} \\
      \midrule
      \name{}-B & \textbf{52.4} & \textbf{57.8} & \textbf{63.4} & \textbf{54.8} & \textbf{59.7} & \underline{62.5} & \textbf{29.5} & \textbf{34.6} & \textbf{37.9} & \textbf{42.9} & \underline{44.8} & 46.0 & \textbf{0.437} & \textbf{0.430} & \textbf{0.429} & \textbf{41.0} & \textbf{45.3} & \underline{49.1} & \textbf{47.8} & \textbf{50.5} & \textbf{53.5} \\
      \midrule
      CanViT (released) & \underline{49.2} & \underline{55.0} & 58.3 & \underline{53.6} & \underline{57.9} & 61.2 & \underline{26.9} & \underline{31.2} & 34.4 & 39.9 & 44.3 & 45.6 & 0.499 & 0.482 & 0.492 & \underline{39.8} & \underline{45.0} & 46.8 & \underline{43.9} & 46.8 & 49.4 \\
      CanViT (our training, released init) & 47.3 & 54.6 & \underline{59.4} & 52.6 & 57.7 & \textbf{62.6} & 26.2 & 30.8 & 35.5 & 39.3 & \underline{44.8} & \underline{46.7} & 0.484 & 0.456 & 0.479 & 38.6 & 43.8 & 48.1 & \underline{43.9} & \underline{47.0} & 50.5 \\
      CanViT (our training, DINOv3 init) & 41.6 & 52.0 & 59.1 & 48.7 & 56.9 & 62.3 & 23.9 & 30.6 & \underline{36.5} & \underline{40.3} & \textbf{45.2} & \textbf{46.9} & \underline{0.469} & \underline{0.435} & \underline{0.465} & 34.6 & 44.0 & \textbf{50.7} & 42.1 & 46.0 & \underline{50.9} \\
      \bottomrule
    \end{tabular}
  \end{adjustbox}
\end{table*}

\begin{table*}[t]
  \centering
  \setlength{\abovecaptionskip}{0pt}
  \caption{\name{}s (finetuned) with each of five teachers, strategy Obj-Tile\textsuperscript{T}. In parentheses: the share of the teacher's value.}
  \label{tab:fovit_teachers}
  \vspace{4pt}
  \footnotesize
  \setlength{\tabcolsep}{2.5pt}
  \begin{adjustbox}{max width=\linewidth}
    \begin{tabular}{@{}l@{\hspace{9pt}}w{c}{4.96em}@{\hspace{9pt}}*{3}{w{c}{4.21em}}@{\hspace{9pt}}w{c}{4.96em}@{\hspace{9pt}}*{3}{w{c}{4.21em}}@{}}
      \toprule
       & \multicolumn{4}{@{}c@{\hspace{9pt}}}{SPair-71k (PCK $\uparrow$)} & \multicolumn{4}{@{}c@{}}{NAVI (Recall $\uparrow$)} \\
      \cmidrule(r{9pt}){2-5} \cmidrule{6-9}
      Teacher & \textit{whole scene} & 1 & 4 & 16 & \textit{whole scene} & 1 & 4 & 16 \\
      \midrule
      DINOv3-B & 54.1 & 41.0 {\scriptsize (76\%)} & 45.3 {\scriptsize (84\%)} & 49.1 {\scriptsize (91\%)} & 57.0 & 47.8 {\scriptsize (84\%)} & 50.5 {\scriptsize (88\%)} & 53.5 {\scriptsize (94\%)} \\
      DINOv2-B & 57.8 & 46.1 {\scriptsize (80\%)} & 50.9 {\scriptsize (88\%)} & 54.8 {\scriptsize (95\%)} & 56.0 & 47.9 {\scriptsize (86\%)} & 51.2 {\scriptsize (91\%)} & 53.8 {\scriptsize (96\%)} \\
      PE-Spatial-B & 49.4 & 38.3 {\scriptsize (78\%)} & 42.2 {\scriptsize (85\%)} & 45.8 {\scriptsize (93\%)} & 50.2 & 43.0 {\scriptsize (86\%)} & 45.4 {\scriptsize (90\%)} & 48.2 {\scriptsize (96\%)} \\
      C-RADIOv3-B & 50.3 & 40.2 {\scriptsize (80\%)} & 44.3 {\scriptsize (88\%)} & 48.0 {\scriptsize (95\%)} & 52.8 & 44.8 {\scriptsize (85\%)} & 48.0 {\scriptsize (91\%)} & 51.3 {\scriptsize (97\%)} \\
      iBOT-B & 43.2 & 37.0 {\scriptsize (86\%)} & 39.4 {\scriptsize (91\%)} & 41.5 {\scriptsize (96\%)} & 48.4 & 43.0 {\scriptsize (89\%)} & 44.6 {\scriptsize (92\%)} & 46.5 {\scriptsize (96\%)} \\
      \bottomrule
    \end{tabular}
  \end{adjustbox}
\end{table*}

\end{document}